\documentclass[journal]{IEEEtran}
\usepackage{algorithm}
\usepackage{algorithmic}
\usepackage{graphicx}
\usepackage{epstopdf}
\usepackage{cite}
\usepackage{makecell}
\usepackage{array}
\usepackage{multirow}
\usepackage{amsfonts,amssymb}
\usepackage{amsmath}
\usepackage{arydshln} 
\usepackage{flushend}
\usepackage{booktabs}
\usepackage{color}
\usepackage{xcolor}
\usepackage{xpatch}
\usepackage{textcomp}

\usepackage{algorithm}
\usepackage{algorithmic}
\makeatletter
\def\changeBibColor#1{%
	\in@{#1}{}%  list of colored bib items cmc, Goyal2017, 5653650
	\ifin@\color{red}\else\normalcolor\fi
}

\xpatchcmd\@bibitem
{\item}
{\changeBibColor{#1}\item}
{}{\fail}

\xpatchcmd\@lbibitem
{\item}
{\changeBibColor{#2}\item}
{}{\fail}
\makeatother

\begin{document}
\newcolumntype{R}[1]{>{\raggedleft\arraybackslash}p{#1}}
\newcolumntype{L}[1]{>{\raggedright\arraybackslash}p{#1}}
\newcolumntype{C}[1]{>{\centering\arraybackslash}p{#1}}

\title{Unsupervised Deep Representation Learning and Few-Shot Classification of PolSAR Images}
\author{Lamei~Zhang,~\IEEEmembership{Senior Member,~IEEE},
	Siyu~Zhang,
	Bin~Zou,~\IEEEmembership{Senior Member,~IEEE}, and 
	Hongwei~Dong
	% <-this % stops a space
	\thanks{This work was supported in part by the National Natural Science Foundation of China (61871158), in part by Aeronautical Science Foundation of China (20182077008, 2018ZC07009). (\emph{Corresponding author: Hongwei Dong; Bin Zou.})}
	\thanks{The authors are with the Department of Information Engineering, Harbin Institute of Technology, Harbin 150001, China (e-mail: donghongwei1994@163.com; zoubin@hit.edu.cn).}}
\markboth{}%IEEE TRANSACTIONS ON GEOSCIENCE AND REMOTE SENSING%
{Shell \MakeLowercase{\textit{et al.}}: Bare Demo of IEEEtran.cls for IEEE Journals}

% make the title area
\maketitle
\begin{abstract}
Deep learning and convolutional neural networks (CNNs) have made progress in polarimetric synthetic aperture radar (PolSAR) image classification over the past few years. However, a crucial issue has not been addressed, i.e., the requirement of CNNs for abundant labeled samples versus the insufficient human annotations of PolSAR images. It is well-known that following the supervised learning paradigm may lead to the overfitting of training data, and the lack of supervision information of PolSAR images undoubtedly aggravates this problem, which greatly affects the generalization performance of CNN-based classifiers in large-scale applications. To handle this problem, in this paper, learning transferrable representations from unlabeled PolSAR data through convolutional architectures is explored for the first time. Specifically, a PolSAR-tailored contrastive learning network (PCLNet) is proposed for unsupervised deep PolSAR representation learning and few-shot classification. Different from the utilization of optical processing methods, a diversity stimulation mechanism is constructed to narrow the application gap between optics and PolSAR. Beyond the conventional supervised methods, PCLNet develops an unsupervised pre-training phase based on the proxy objective of instance discrimination to learn useful representations from unlabeled PolSAR data. The acquired representations are transferred to the downstream task, i.e., few-shot PolSAR classification. Experiments on two widely-used PolSAR benchmark datasets confirm the validity of PCLNet. Besides, this work may enlighten how to efficiently utilize the massive unlabeled PolSAR data to alleviate the greedy demands of CNN-based methods for human annotations.
\end{abstract}
\IEEEpeerreviewmaketitle
\section{Introduction}
\IEEEPARstart{P} {olarimetric} synthetic aperture radar (PolSAR) image classification aims to predict each pixel of the whole map. It has been a hot topic because of the powerful observation capacity of PolSAR system. The development of many industries, such as agriculture \cite{application2}, urban planning \cite{application3}, geoscience \cite{application1}, environmental monitoring \cite{application4,application5}, etc., is inseparable from the valuable information extracted by PolSAR classification. Therefore, the significance of the breakthrough of PolSAR classification is not limited to itself, but also lies in the broad application fields.
\par Deep learning, represented by convolutional neural networks (CNNs) \cite{Lecun2014Backpropagation}, has made progress in many problems, e.g., optical \cite{Krizhevsky2012ImageNet,Szegedy2014Going,ResNet}, medical \cite{UNet,7463094,7404017} and remote sensing \cite{MA2019166,8113128,7486259} image recognition. Due to the impressive results achieved by CNNs, the mainstream feature extraction technique of PolSAR classification is currently transforming from unsupervised hand-crafted features with physical meanings \cite{oldmethod3,oldmethod2,wishart,srw} to supervised deep ones obtained by neural networks. Zhou \emph{et al.} firstly explored the application of CNNs in PolSAR image classification \cite{7762055}. They constructed a four-layer convolutional architecture to process the 6-D manually designed PolSAR representations, and the experiments showed breakthrough results. Recently, the nonlinear fitting ability of CNNs has attracted widespread attention, and various supervised CNN-based PolSAR studies are springing up. Some focused on how to find suitable input information to boost the classification performance, such as manually \cite{input1} or auto-selected \cite{8771134} polarimetric features, raw complex-valued PolSAR data \cite{complex} or the improved versions \cite{input2,input3}. Besides, many studies concerned about using advanced CNN models, such as fully convolutional \cite{8936481}, 3D convolution-based \cite{8864110}, sparse manifold-regularized \cite{8835090}, generative \cite{8765386} and hyperparameter optimized \cite{9034477} architectures. 
\par The recently developed supervised CNN-based methods have achieved promising results and improved PolSAR classification to some extent \cite{8899902}. But this does not mean that unsupervised methods are no longer needed; on the contrary, their existences become more essential. The supervised machine learning paradigm implies that the high recognition accuracy is based on a sufficiently large training set with human annotations \cite{PCA}, especially for deep CNNs with a large number of trainable parameters. The intrinsic reason may be that the training process based on sparse labels is easy to converge to a fragile and  task-specific solution \cite{sslsurvey}. Although augmentation and regularization techniques \cite{Srivastava2014Dropout, ReLU, highwaynetwork} were explored, this requirement is still hard to meet in the application of easily acquired and understood optical images, let alone the more complex PolSAR systems. Insufficient supervision will cause the network to overfit the training data, thus lacking generalization in large-scale applications, which can be regarded as the most significant bottleneck hindering CNNs-based PolSAR classifiers. Therefore, unsupervised CNNs which combine the advantages of both, i.e., the discrimination ability of CNNs and the feasibility for large-scale problems of unsupervised methods, are undoubtedly more desirable and meaningful than supervised ones.
\par This work falls in the area of unsupervised PolSAR representation learning \cite{6472238}. Similar to the supervised, unsupervised methods can be implemented by shallow models and deep neural networks. The former is widely-used in PolSAR area, including a variety of physical \cite{freeman,halpha} and statistical \cite{super-wishart} features. The complexity of these methods is low, which brings fast running speed but also limits the performance. In contrast, neural networks for unsupervised learning are highly flexible and effective. Autoencoder is a representative technique, which explicitly defines a feature extraction mapping through the objective of image reconstruction \cite{NIPS1993_798}. The features learnt by encoders can be used by network fine-tuning  \cite{8356045,7518668}. Although it is hard to evaluate the quality of such automated feature engineering, some recent studies showed that the reconstruction loss based methods may be difficult to learn high-level representations because they pay too much attention to pixel-level details. % The limitation of reconstruction also leads to a wide range of thinking: What kind of objective is suitable for learning high-level representations without human supervision? 
\par Greatly inspired by the success in natural language processing \cite{bert}, self-supervised learning (SSL) \cite{DoerschContextPrediction} provides a promising way for unsupervised representation learning, which follows the supervised learning paradigm, but the supervision is provided by the data itself. Therefore, free and abundant labeled samples are available for network training due to the automation of the pseudo-label generation process. Moreover, self-generated annotations can provide richer information, because the pale human-made labels cannot indicate the potential connections between samples of the related categories. Similar to some few-shot learning paradigms \cite{maml,CACTUS,matching}, SSL can be divided into two components, i.e., unsupervised pre-training and classifier fine-tuning. Designing effective pre-training methods to acquire transferrable representations is the key to the validity of SSL. Generally, the pre-training is performed by a proxy objective, called pretext task, and the genuine interest (PolSAR classification in this paper) is called downstream task. The construction of most pretext tasks is heuristic and predictive, such as predicting spatial correlations \cite{rotations,Jigsaw} and colors \cite{Colorization}. Although they have achieved some results, the generality of these pretext tasks is obviously not enough \cite{simclr}. For example, it is meaningless to predict the spatiality for satellite images and the color for CT images. Recently, a flexible paradigm of SSL based on the pretext task of instance discrimination \cite{InstanceDiscrimination} and InfoNCE loss function \cite{cpc,deepinfo,amdim}, i.e. contrastive learning (CL), has emerged and made a breakthrough \cite{moco}. The proposal of instance discrimination comes from the fact that the apparent similarity among semantic categories can be automatically discovered by neural networks. Therefore, the similarity among instances may also be captured, which can be used as high-level representations. InfoNCE loss is a good way to cooperate with the instance discrimination, which serves as a measure to maximize the mutual information between instances \cite{deepinfo}.
\begin{figure}[h]
	\begin{centering}
		\includegraphics[width=9.0cm]{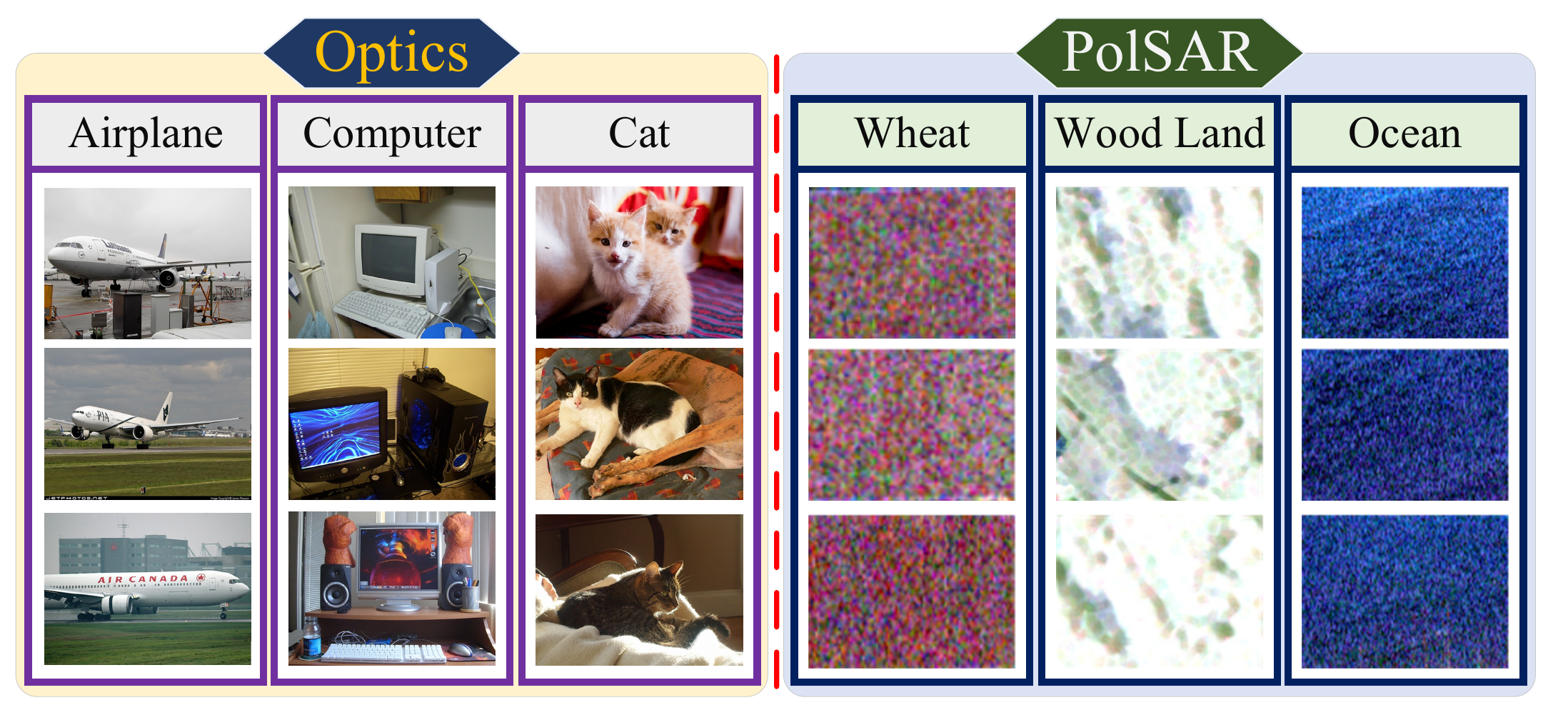}
		\caption{An intuitive comparison about the diversity of individuals between optical images and PolSAR images.}\label{fig:g1}
	\end{centering}
\end{figure} 
\par Considering the appealing properties of CL, the objective of this work is to combine it with PolSAR images and achieve high-precision few-shot classification \cite{fewshot}. However, it must be pointed out that all existing CL methods are proposed for optical image processing. Although the generality is intrinsic, the application gap between optics and PolSAR can not be ignored. The authors believe that CL should be transformed into PolSAR-tailored methods to obtain satisfactory results, rather than following the original blindly. We maintain that the key factor that affects the performance of CL in PolSAR representation learning is not the data modality, but diversity. It can be found that the transferrable representations are learnt through distinguishing the difference between individuals in CL methods. In other words, if the similarity between each sample is high, the performance will be greatly reduced. As shown in Fig. \ref{fig:g1}, for optical images, the difference is relatively large whether they are between different categories (inter-class) or between different instances of the same category (intra-class). But it is another story for PolSAR images. For a pair of optical image and PolSAR image, the size of the corresponding real scene is quite different. This phenomenon is reflected in PolSAR data with the following two characteristics: less number of categories and lower intra-class diversity. And it brings a great challenge to the optimization of CL, i.e., a large number of samples from the same category have to be selected during random sampling, and they are difficult to identify with each other.

\par Based on the above analysis and inspired by previous works, a PolSAR-tailored contrastive learning network (PCLNet) is proposed in this paper. The proposal effectively combines the unsupervised CL methods with PolSAR representation learning and classification. The main novelties and contributions can be summarized as follows:
\begin{enumerate}
	\item Unsupervised deep PolSAR representation learning and few-shot classification are explored with the help of CL for the first time. Specifically, an unsupervised pre-training method is designed to learn transferrable representations without human annotations. The acquired representations are transferred with very little supervision to achieve few-shot PolSAR classification. Theoretically speaking, we construct a practical way to utilize the massive unlabeled PolSAR data and improve the applicability of CNN-based methods to large-scale problems.
	\item A novel diversity stimulation mechanism is proposed and combined with the CL method, which narrows the application gap between optics and PolSAR. The diversity of training samples can be stimulated through two steps: Firstly, revised Wishart distance \cite{srw} based unsupervised clustering is used to perform an overcomplete partition of the dataset and construct numerous categorizations. Then, fully connected graphs are constructed for each category, and the nodes with high affinities are removed. The diversified training data acquired by this dual-stimulating mechanism can be seen as the key factor that makes CL methods work in few-shot PolSAR classification.
	\end{enumerate}
\par The rest of this paper is organized as follows: The proposed PCLNet is introduced in Section \ref{sec:2}. Experimental results and analyses are presented in Section \ref{sec:exp}. Section \ref{sec:con} concludes this work and gives possible future directions.

\section{Proposed method}\label{sec:2}
\begin{figure}[h]
	\begin{centering}
		\includegraphics[width=8.5cm]{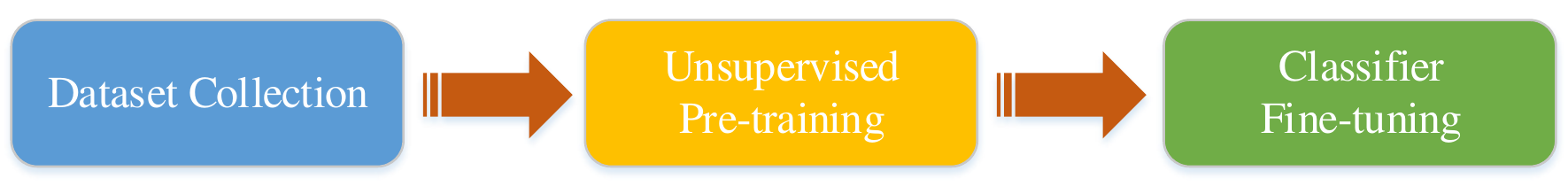}
		\caption{General flow chart of the training of PCLNet.}\label{fig:0}
	\end{centering}
\end{figure}

In this section, the proposed PCLNet for unsupervised deep PolSAR representation learning and few-shot classification is presented. The proposal is a variant of CL, which is customized for PolSAR images. As shown in Fig. \ref{fig:0}, the training of PCLNet includes three steps. Among them, the first step is the foundation, which supports the subsequent programs. The middle step is the most important one, which obtains deep PolSAR representations without supervision. To accommodate specific classification tasks, in the third step, a classifier will be fine-tuned with a very small amount of supervision.	
\par Since there is no benchmark for CL, we first need to construct a dataset for unsupervised learning. It is worth noting that in this process, manually labeling is unnecessary, which supports the use of massive unlabeled PolSAR data. The acquired dataset is then used for unsupervised network pre-training. Finally, a new round of network fine-tuning is performed on basis of the result of unsupervised pre-training. In fact, the training of traditional CNN-based PolSAR classifiers can be regarded as doing the third step from scratch. The following is a detailed description of each step. 
\subsection{Dataset Collection}
The construction of training dataset is the first problem to be solved. Random sampling is a natural choice used by almost all CL methods, which is to randomly select a certain number of samples to form the training set. However, this method is not suitable for PolSAR images. Because the validity of random sampling is supported by the differences between individuals in the dataset, and such differences are obviously much smaller in PolSAR images. Moreover, the difference between the data obtained by various PolSAR systems is relatively large. Therefore, an applicable dataset collection method is more desired rather than specific datasets. To address these issues, a diversity stimulation mechanism is designed as a generic means to obtain the dataset with a high degree of diversity for the training of CL. This is a dual mechanism, which is realized by successively stimulating the inter-class and intra-class diversities.
\subsubsection{Stimulation for Inter-Class Diversity}
A widely-used clustering method, i.e., unsupervised Wishart classifier, is adopted to perform a preliminary overcomplete partition for PolSAR images. The Wishart classifier is based on central grouping techniques and inherits many attractive highlights of the well-known K-means algorithm \cite{kmeans}. 
\par According to the basic operation principle of PolSAR \cite{scatter}, the complex Sinclair scattering matrix $\mathbf{S}$ is usually utilized to represent the amplitude and phase information of the transmitted and received backscattered signals. In a dynamically changing environment, numerous distributed targets can be analyzed by the polarimetric coherency matrix $\mathbf{T}$ which follows complex Wishart distribution:
\begin{equation}
\mathbf{T}=\frac{1}{n} \sum_{i=1}^{n} \underline{k}_{i} \cdot \underline{k}_{i}^{H}
\end{equation}
where $H$ denotes the complex Hermitian transpose, $n$ is the number of looks and $\underline{k}$ is the Pauli scattering target vector. Based on the matrix Wishart distance, Lee \emph{et al.} \cite{wishart} introduced the unsupervised Wishart classifier to assign each pixel of coherency matrix with a cluster prototype $\hat{\mathbf{V}}_{i}$, $i=1, \ldots, K$ where $K$ is the number of clusters. For example, if one pixel is corresponding to class $\omega_{m}, m \in\{1, \ldots, K\}$, then
\begin{equation}
d_{W}\left(\mathbf{T}, \hat{\mathbf{V}}_{m}\right) \leq d_{W}\left(\mathbf{T}, \hat{\mathbf{V}}_{i}\right), \forall \omega_{i} \neq \omega_{m}.\label{2}
\end{equation}
\par Considering that the revised Wishart distance \cite{srw} satisfies the identity of discernibles $d_{W}(\mathbf{T}, \mathbf{T})=0$ and symmetry conditions $d_{W}(\mathbf{T}, \hat{\mathbf{V}}_{m})=d_{W}(\hat{\mathbf{V}}_{m}, \mathbf{T})$, it is used to measure the pair-wise distance between samples and cluster prototypes:
\begin{equation}
d_{W}\left(\mathbf{T}, \hat{\mathbf{V}}_{m}\right)=\frac{1}{2} \operatorname{tr}\left(\mathbf{T} \hat{\mathbf{V}}_{m}^{-1}+\hat{\mathbf{V}}_{m}  \mathbf{\mathbf{T}}^{-1}\right)-r \label{1}
\end{equation}
where $\operatorname{tr}(\cdot)$ is the trace of a matrix and $r$ notes the dimension of coherency matrix.
 \begin{figure}[h]
	\begin{centering}
		\includegraphics[width=8.5cm]{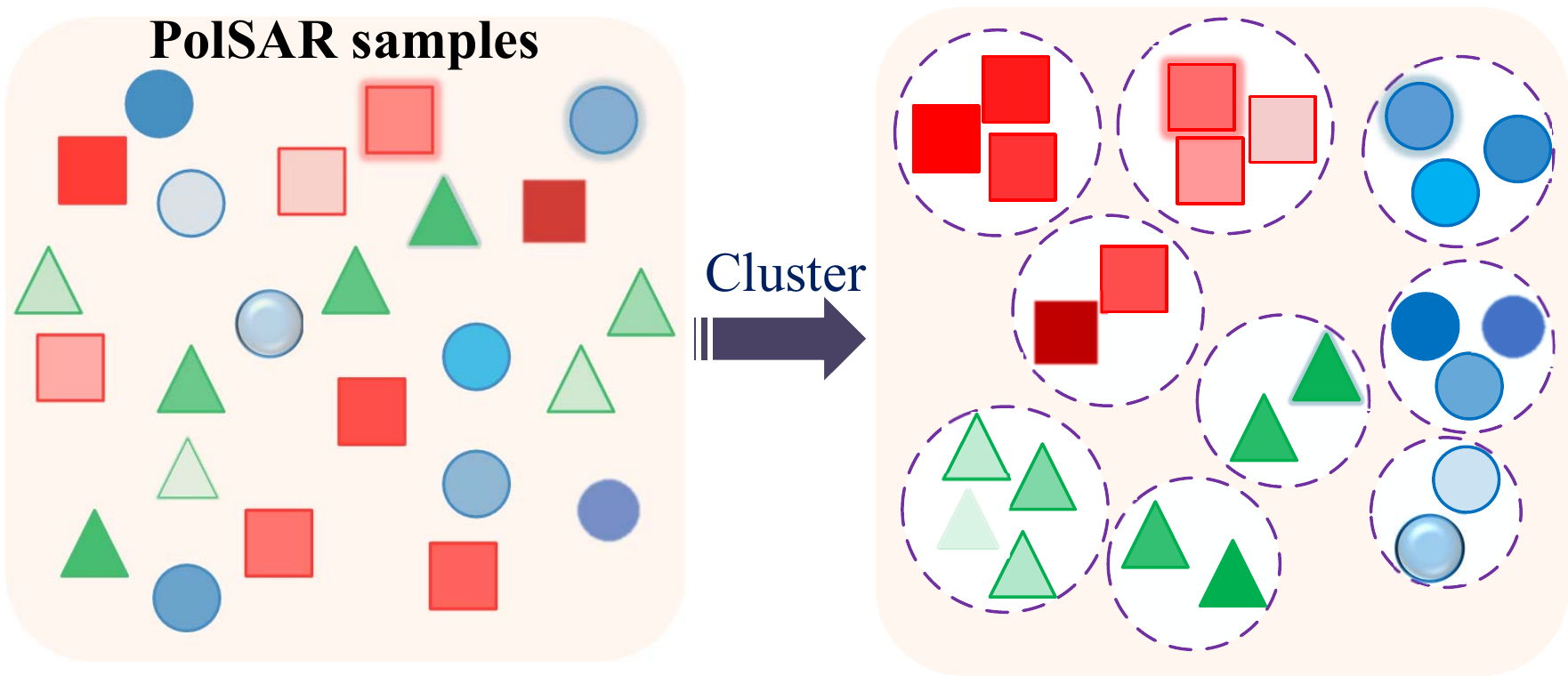}
		\caption{Illustration of the unsupervised clustering for inter-class diversity stimulation. The unlabeled PolSAR data are overcompletely partitioned according to their similarities between each other without human annotations.}\label{fig:g2}
	\end{centering}
\end{figure}
\par As shown in Fig. \ref{fig:g2}, unlabeled PolSAR image samples can be clustered by the revised Wishart distance based unsupervised Wishart classifier. In our setting, the result of clustering should be overcomplete, which means that the number of clusters is unrealistically large. The purpose is to constrain the diversity of clustering prototypes deliberately, so as to stimulate the inter-class diversity.

\subsubsection{Stimulation for Intra-Class Diversity}
Following the result of unsupervised clustering, the training set can be collected through intra-class screening, which is to maintain relatively large diversity between different instances in the same cluster. For the $i$th cluster $\omega_{i}$ with the samples of $\Theta_{i}=\left\{\mathbf{T}_{1}, \ldots, \mathbf{T}_{N_i}\right\}$, an undirected fully connected graph $\mathcal{G}_i=\langle\Theta_i, \mathcal{E}_i\rangle$ can be constructed according to the spectral graph theory \cite{srw}. Then the affinity between any two nodes of $\mathcal{G}_i$ can be represented by the edge $\mathcal{E}_i$. It is worth noting that the nodes of the graph represent the corresponding instances in the cluster, and the affinity can be seen as a tool for measuring the pair-wise similarity between two instances. Hence, the training dataset for unsupervised pre-training can be flexibly collected by cutting each graph.
\par The graph $\mathcal{G}_i$ can be obtained by calculating the affinity between every two instances. In this work, the affinity is defined by a Gaussian kernel function based revised Wishart distance, which can be expressed as:
%\frac{1}{2} \operatorname{tr}\left(\mathbf{T} \hat{\mathbf{V}}_{m}^{-1}+\hat{\mathbf{V}}_{m} ^{-1} \mathbf{\mathbf{T}}\right)-q
\begin{equation}
\begin{split}
\mathcal{A}_i\left(\mathbf{T}_{p}, \mathbf{T}_{q}\right)&= \exp \left\{-\frac{d_{W}^{2}\left(\mathbf{T}_{p}, \mathbf{T}_{q}\right)}{2 \gamma^{2}}\right\}\\
&=\exp \left\{-\frac{\left(\frac{1}{2} \operatorname{tr}\left(\mathbf{T}_{p} \mathbf{T}_{q}^{-1}+\mathbf{T}_{q}  \mathbf{T}_{p}^{-1}\right)-r\right)^{2}  }{2 \gamma^{2}}\right\}\label{3}
\end{split}
\end{equation}
where $\mathcal{A}_i\left(\mathbf{T}_{p}, \mathbf{T}_{q}\right)$ notes the affinity between the instances $\mathbf{T}_{p},\mathbf{T}_{q}$ of $\Theta_{i}$ $(p \neq q)$. $\gamma$ is the Gaussian kernel bandwidth. It is obvious that the graph $\mathcal{G}_i$ is a symmetric positive semidefinite matrix with the size of $N_{i} \times N_{i}$ and the value of its diagonal elements is set to be one. Therefore, only the upper triangular elements need to be calculated.
\begin{figure}[h!]
	\begin{centering}
		\includegraphics[width=8.5cm]{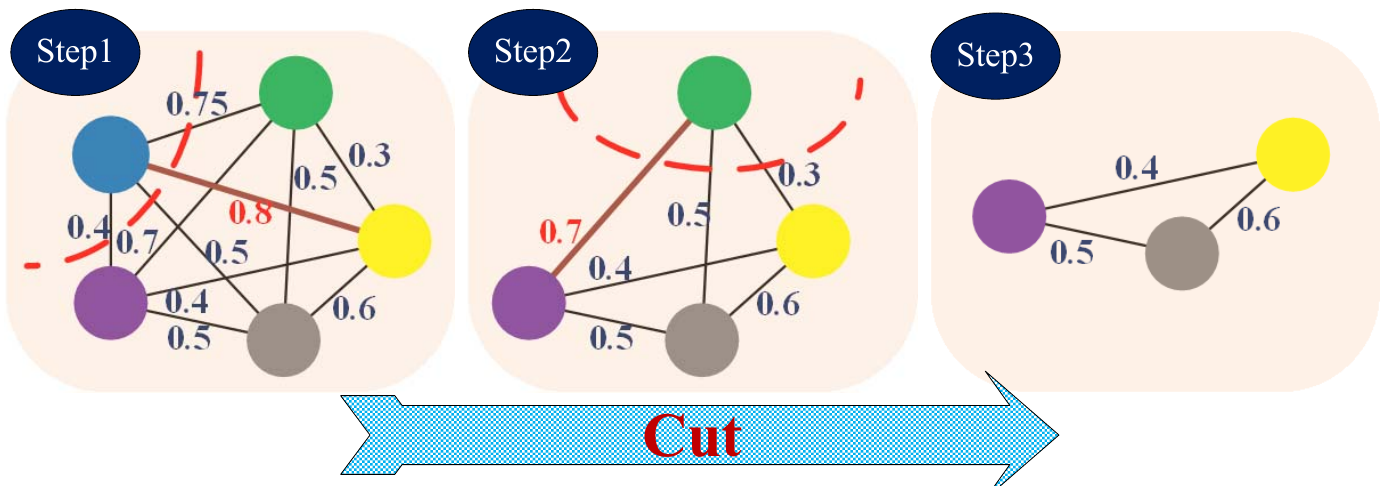}
		\caption{A sketch map for intra-class diversity stimulation. Considering that there are five samples in the cluster, and three of them will eventually be used for the unsupervised pre-training. The fully connected graph is visualized where the circles in different colors represent samples, and values on the lines represent affinities. The edge with the highest affinity is marked in brown. The red dotted line indicates that the sample is diametrically removed.}\label{fig:g3}
	\end{centering}
\end{figure}
\par As shown in Fig. \ref{fig:g3}, intra-class diversity stimulation of each cluster is implemented by cutting the nodes (samples) of the corresponding fully connected graph. First, all of the upper triangular elements in the fully connected graph are sorted, and the affinity with the largest value is located. Next, one of the two nodes connected by this edge will be randomly removed. This process will be carried out iteratively until the number of samples reaches a pre-defined threshold. Finally, the instances corresponding to the remaining nodes will be collected to form the dataset for the training of CL. The process of collecting a dataset by the proposed diversity stimulation mechanism is outlined as Algorithm \ref{alg:1}. 
\begin{algorithm}[h!]
	\caption{Diversity Stimulation Mechanism for Dataset Collection}
	\label{alg:1}
	\begin{algorithmic}[1]
		\STATE \textbf{Begin}
		\STATE Prepare PolSAR image samples $\Theta=\left\{\mathbf{T}_1, \mathbf{T}_2, \ldots, \mathbf{T}_N\right\}$ in the form of polarimetric coherency matrix. \\
		\REQUIRE Inter-class diversity stimulation\\
		\STATE Prepare the number of clusters $K$, max iterations $\mathcal{I}$ and hyperparameter $r$. Initialize cluster prototypes $\hat{\mathbf{V}}_{i}, i=1, \ldots, K$. \\
		\FOR{iteration in $\mathcal{I}$:}
		\STATE Measure the revised Wishart distance $d_W$ between samples and cluster prototypes by \eqref{1}.\\
		\STATE Assign each sample with corresponding cluster by \eqref{2}.\\
		\STATE Recalculate cluster prototypes $\hat{\mathbf{V}}_{i}, i=1, \ldots, K$.\\
		\ENDFOR
		\RETURN cluster sample sets $\Theta_{i}=\left\{\mathbf{T}_{1}, \mathbf{T}_{2}, \ldots, \mathbf{T}_{N_i}\right\}$, $i=1, \ldots, K$.\\
		\ENSURE Intra-class diversity stimulation\\
		\STATE Prepare the number of retaining sample threshold $M$ and hyperparameter $\gamma$. \\
		\FOR{cluster in $K$:}		
		\STATE Construct the fully connected graph by \eqref{3} for all samples in the cluster.\\
		\WHILE{$N_i>M$:}
		\STATE $(p^{*}, q^{*})=\mathop{\arg\max}_{p,q} \{ \mathcal{A}\left(\mathbf{T}_{p}, \mathbf{T}_{q}\right)\}$, randomly remove one sample $\mathbf{T}_{p^{*}}$ or $\mathbf{T}_{q^{*}}$ from $\Theta_{i}$. \\
		\ENDWHILE
		\STATE Add the retaining samples in the cluster to dataset $\Theta$.\\
		\ENDFOR
		\RETURN dataset $\Theta$
	\end{algorithmic}
\end{algorithm}

\subsection{Unsupervised Pre-training}
\par Inspired by some related works \cite{moco, simclr, InstanceDiscrimination, amdim, deepinfo, cpc}, a CL based unsupervised pre-training method is designed in this part. The novelty is that the training of the proposal can be implemented without human annotations, which brings the possibility of using massive unlabeled PolSAR images. Moreover, the transferrable deep PolSAR representations can be acquired by the pre-trained network, which are the bases for achieving few-shot classification. It is worth noting that the implementation of unsupervised pre-training is supported by the diversified training data obtained by the above dual-stimulating mechanism. The following points need to be considered during the construction of unsupervised pre-training: pretext task and loss function, the architecture of encoder and its optimization. 
\subsubsection{Pretext Task and Loss Function}
\par Generally speaking, high-level representations work better when transferring to other tasks because they are more abstract than low-level ones. It has been proved that the training of supervised learning is inefficient and it converges to a fragile and task-specific solution \cite{sslsurvey,8633866}. In other words, although the representations obtained by supervised CNNs are higher-order than hand-crafted ones, they are still not robust enough to achieve the task migration. 
\par To address this issue, the objective and corresponding loss function of supervised learning should be improved. As known that the training of most supervised methods is based on category discrimination, so it is necessary to provide category information manually. In this work, instance discrimination \cite{InstanceDiscrimination} is adopted which takes the category-wise supervision to the extreme, i.e., treat each sample as a category. Therefore, the sample itself provides the supervision and human annotation is no longer needed. The validity of such a pretext task comes from the inference that realizing instance discrimination requires more generalized representations than category discrimination. 
\par After the objective of instance discrimination is determined, an instance-wise metric needs to be established instead of the utilization of cross-entropy loss. As the currently most widely-used contrastive loss function, InfoNCE loss \cite{cpc, cmc, simclr} is used in this work to implement instance discrimination by maximizing the mutual information \cite{amdim, deepinfo}. Considering that there are $M$ samples to form a training set $O=\left\{o_1, o_2, \ldots, o_M\right\}$, and another view of $O$ can be expressed as $O'=\left\{o'_1, o'_2, \ldots, o'_M\right\}$ where each $o'_i$ corresponds to the $o_i$ in the original training set. Then the InfoNCE loss for $o_i$ can be defined as:		
%\begin{small}
\begin{equation}
\mathcal{L}(o_i, o_i^{+}, o_j^{-})\!=\!-\!\log\! \frac{\exp \left(\left\langle{o_i, o_i^{+}} \right\rangle\! /\! \tau\right)}{\exp \left(\left\langle{o_i, o_i^{+}} \right\rangle \!/\! \tau\right)\!+\!\sum_{j} \exp \left(\left\langle{o_i, o_j^{-}} \right\rangle\! / \!\tau\!\right)}\label{00}
\end{equation}
%\end{small}
where $o_i^{+}$ denotes the element related to $o_i$ in $O'$, i.e., $o'_i$. $o_j^{-}$ denotes any element in set $O'$ except $o_i^{+}$, and the upper limit of $j$ should be $M-1$. $\langle\cdot, \cdot\rangle$ represents the cosine similarity \cite{cosine} and $\tau$ is the temperature \cite{InstanceDiscrimination} hyperparameter that controls the uniformity of information distribution. In most CL based studies, $o_i$ and $o_i^{+}$ are treated as a positive pair, and $o_i$ with all of $o_j^{-}$ are treated as negative pairs. As described in the definition, \eqref{00} can be seen as a multi-class $M$-pair loss \cite{npairloss} which tries to classify $o_i$ as $o_i^{+}$.
\subsubsection{Architecture of Encoder}
\par
\begin{figure}[ht]
	\begin{centering}
		\includegraphics[width=8.5cm]{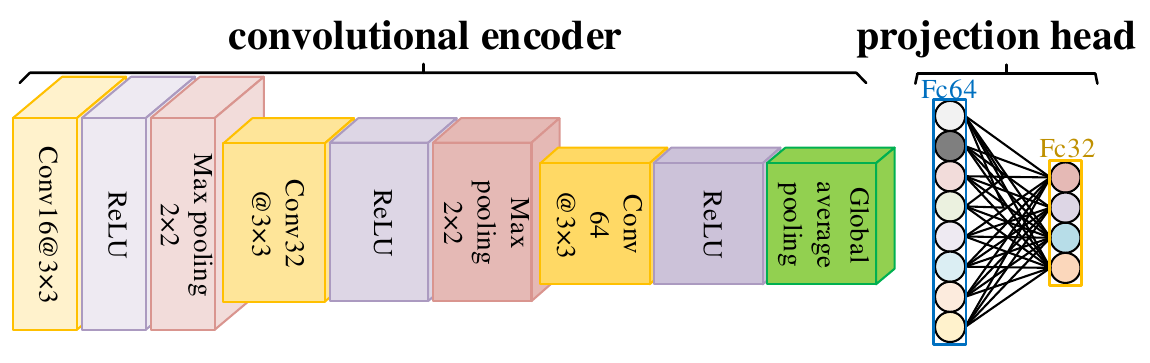}
		\caption{Architecture of the encoder for obtaining deep PolSAR representations in PCLNet.}\label{fig:g4}
	\end{centering}
\end{figure}

 In PCLNet, an encoder is used to obtain the representations of input samples. The encoder can be divided into convolutional encoder and projection head, and the former is what we want to obtain through the unsupervised pre-training. The convolutional encoder consists of four parts, including convolution, nonlinear activation, pooling, and global average pooling. An intuitive diagram of the encoder is shown in Fig. \ref{fig:g4}. 
\par The calculation of convolution can be defined as:
\begin{equation}
v_{i}^{(l+1)}=\sum_{d=1}^{D} w_{i d}^{(l+1)} * x_{d}^{(l)}+b_{i}^{(l+1)}
\end{equation}
where $x_{d}^{(l)}$ and $v_{i}^{(l+1)}$ represent the $d$th input and the $i$th output of layer $l+1$, $w$ and $b$ denote the learnable kernel matrix and bias, $D$ is the number of input feature maps and $*$ denotes the convolution operator. To improve the nonlinear ability and avoid gradient vanishing, rectified linear units (ReLU) \cite{ReLU} is employed as the activation function, which is implemented by:
\begin{equation}
\sigma_{\mathrm{ReLU}}\left(x\right)=\max \left(0,x\right)
\end{equation}
\par Pooling can be considered as a tool for sub-sampling, which reduces the dimension of features. Moreover, it helps to identify displacement, scaling, and other distortion-invariants in 2D maps. After building several network layers in the form of conv-ReLU-maxpool, global average pooling (GAP) is employed to convert the obtained feature maps into feature vectors by computing the mean of the height and width of the feature maps \cite{linNIN}. This operation can also decrease the computational complexity effectively.

\par Note the aforementioned convolutional encoder as $f_{\theta}(\cdot)$ where $\theta$ means all the learnable parameters, a PolSAR image sample $\mathbf{x}_{i}$ can be transformed into a representation $h_i=f_{\theta}\left(\mathbf{x}_{i}\right)$. Then a projection head is used to map the representation into the space where contrastive loss is applied. Some recent studies have shown that such a module can prevent the loss of information valid for the downstream tasks \cite{moco,simclr}. In this work, a multilayer perceptron is adopted to construct the projection head. Note the projection head as $g_{\varphi}(\cdot)$ where $\varphi$ means its learnable parameters, for the input $\mathbf{x}_{i}$, the output of encoder can be written as:
\begin{equation}
o_{i}=g_{\varphi}\left(h_{i}\right)=\mathbf{W}^{\left(2\right)}\sigma_{\text{ReLU}}\left(\mathbf{W}^{\left(1\right)}f_{\theta}\left(\mathbf{x}_{i}\right)+\mathbf{b}^{\left(1\right)}\right)+\mathbf{b}^{\left(2\right)}\label{08}
\end{equation}
where $o_i$ denotes the deep representation of PolSAR sample $\mathbf{x}_{i}$. $\mathbf{W}, \mathbf{b}\in \varphi$ denote the weight matrix and bias of the projection head. We name the encoder composed of $f_{\theta}(\cdot)$ and $g_{\varphi}(\cdot)$ as main encoder, which is used to obtain deep representations from input samples. 
\par Although not shown in Fig. \ref{fig:g4}, the encoder of PCLNet is actually a two-stream architecture \cite{twostream} and the two branches share the topology and hyperparameter. In other words, there is another architecture, called auxiliary encoder, which exists in parallel with the main encoder. Such design is determined by the definition of InfoNCE loss. It can be found that pairs of positive and negative samples are needed for the calculation of InfoNCE loss. To obtain them, correlated views of PolSAR image samples $\mathbf{x}$ should be generated through data augmentation methods. Specifically, as a view or augment of the $i$th sample, $\mathbf{x}_i^{+}$ will be generated and combined with $\mathbf{x}_i$ to form an original positive pair. Similar to \eqref{08}, fed $\mathbf{x}_i^{+}$ to the auxiliary encoder and the output can be written as: 
\begin{equation}
o_{i}^{+}=g_{\tilde{\varphi}}\left(f_{\tilde{\theta}}(\mathbf{x}_i^{+})\right)
\end{equation}
where $f_{\tilde{\theta}}(\cdot)$ and $g_{\tilde{\varphi}}(\cdot)$ denote the convolutional encoder and projection head of the auxiliary encoder, respectively. Positive pairs, i.e., $o$ and $o^{+}$, can be produced by the two branches of the encoder. Since each output of the auxiliary encoder has a dual identity, i.e., for the $i$th output, it is both a positive sample of $o_i$ and a negative sample of all $o_j \ (j\neq i)$, so the calculation of InfoNCE loss can be supported. It is worth noting that the relationship between positive pairs is similar to that of samples and labels in supervised learning. 
\subsubsection{Optimization of Encoder}
\begin{figure*}[t]
	\begin{centering}
		\includegraphics[width=\textwidth]{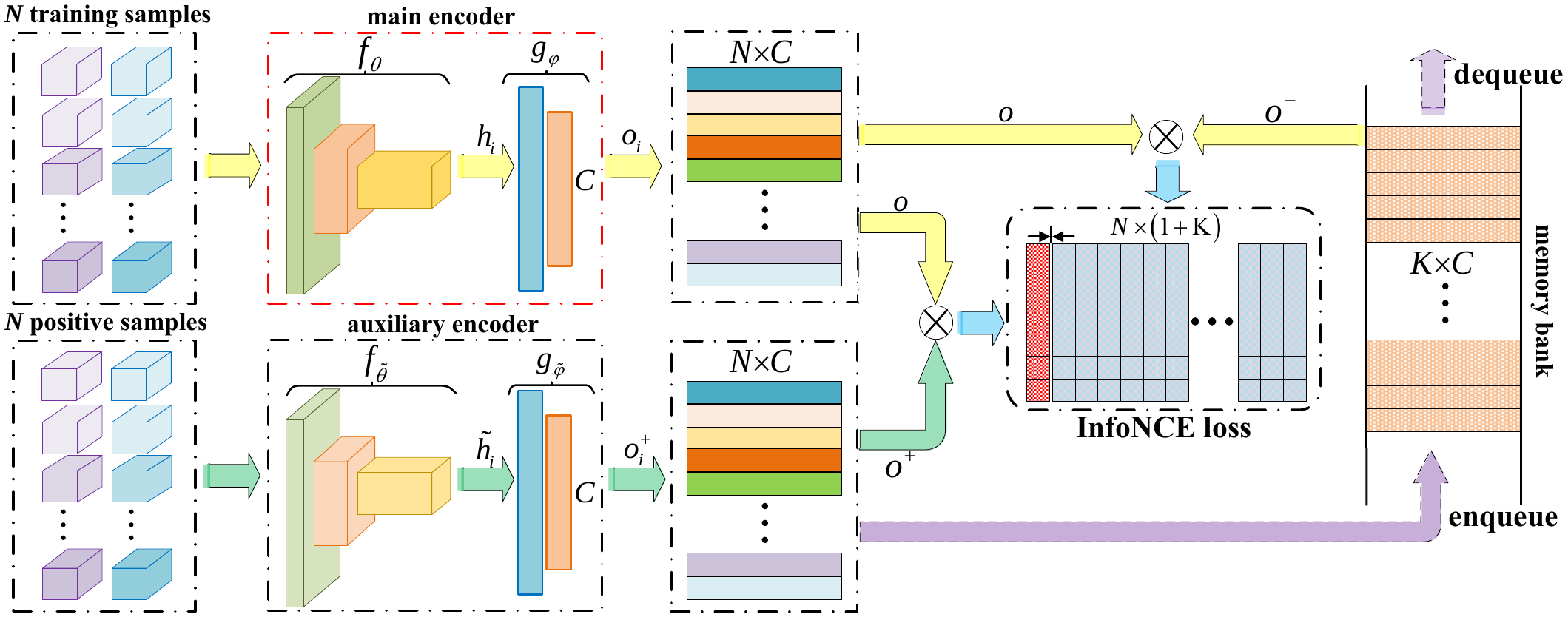}
		\caption{General flow chart of the optimization of main encoder, whose update can be realized through the error backpropagation of the InfoNCE loss. The memory bank stores the output of auxiliary encoder in previous mini-batches. To increase the number of negative sample pairs, the output of auxiliary encoder in the current mini-batch will not participate in the calculation of InfoNCE loss as negative samples, replaced by the ones stored in the memory bank. The whole optimization process gets rid of human annotations.}\label{fig:g5}
	\end{centering}
\end{figure*}
As stated before, the encoder consists of main encoder and auxiliary encoder. In the design of PCLNet, the optimization of them is not the same. The first thing to point out is that the goal of optimization is the learnable parameters of main encoder, because only the representations it produces will be used in the downstream task. Optimization of the main encoder is implemented by error backpropagation of InfoNCE loss via mini-batch stochastic gradient descent (SGD). Consider a mini-batch with $N$ training samples and the dimension of encoded samples is $C$, a sketch map can be seen from Fig. \ref{fig:g5}. It can be found that the InfoNCE loss plays an important role in the optimization process. As stated above, positive and negative pairs support the calculation of InfoNCE loss. However, directly using \eqref{00} may cause serious performance degradation. The reason is that the effectiveness of InfoNCE loss has a premise: The number of negative sample pairs needs to be very large \cite{InstanceDiscrimination}. When using mini-batch SGD to optimize a CL method, the number of negative sample pairs, i.e., the $M$ in \eqref{00}, is actually equal to the mini-batch size. So a very large mini-batch size is needed, but this will bring some adverse impacts. Training with a very large mini-batch size needs rich memory space to store a large number of training samples, which is not supported by most standard computational platforms. Not only that, it also reduces the efficiency of optimization \cite{Goyal2017}. Although the use of high-performance computational platforms can alleviate this problem \cite{simclr}, the algorithm that can be applied to standard computers is undoubtedly more valuable for promotion. Inspired by previous studies \cite{InstanceDiscrimination, cmc, moco}, in this paper, a memory bank is introduced to provide the negative samples used for the calculation of InfoNCE loss.
\par As shown in Fig. \ref{fig:g5}, in the current mini-batch, positive and negative samples are taken from auxiliary encoder and memory bank, respectively. First, correlated views of the $N$ input PolSAR image samples in this mini-batch are generated. Then, they are fed into the main and auxiliary encoders to obtain $N$ positive pairs (bars of the same color in the two black dotted boxes of the third row of Fig. \ref{fig:g5} indicate positive  pairs). Finally, the samples stored in memory bank are taken out as negative samples to support the calculation of InfoNCE loss, so that the main encoder can be updated through backpropagation. Next, we introduce the acquisition of negative samples and the construction of memory bank.
%\subsubsection{Learning with A Memory Bank}
\par On the premise of obtaining positive pairs, it is important to traverse as many negative pairs as possible \cite{deepinfo}. An important observation is that for the encoded samples $o$ in the current mini-batch, the $o^{+}$ output by all other mini-batches can be regarded as their negative samples. Because in one iteration, the samples contained in each mini-batch have no intersection. Intuitively, outputs of the auxiliary encoder in previous mini-batches have the potential to be reused. Therefore, a memory bank \cite{InstanceDiscrimination} that stores the $o^{+}$ output by the previous $k$ mini-batches is constructed in this paper. The capacity of memory bank is $K$ where $K=k\times N$, so the value of $K$ can be much larger than $N$ as long as $k$ takes a larger value. Due to the addition of memory bank, in the current mini-batch, outputs of the auxiliary encoder will only be used as positive samples, and negative samples can be retrieved directly from the memory bank. Such setting decouples the number of negative samples from the mini-batch size, because the $M$ in \eqref{00} is changed to be the memory bank capacity $K$ instead of the mini-batch size $N$. Retrieving data from memory bank does not require additional calculations, so it is possible to efficiently traverse a large number of negative samples.
\par It should be pointed out that individuals of the memory bank in PCLNet are not static, but vary on-the-fly. After the training of each mini-batch, $N$ representations obtained by the auxiliary encoder will be stored in the memory bank. When the storage of memory bank reaches the upper limit, the representations of the current mini-batch will enqueue and replace the ones of the oldest mini-batch so as to achieve the dynamic update. There are two benefits of making the memory bank dynamic. On the one hand, since the auxiliary encoder will be optimized and its output will evolve, using the latest representations and removing the oldest ones can boost the consistency of the individuals in memory bank \cite{moco}. One the other hand, compared with storing all the representations \cite{InstanceDiscrimination}, the dynamic update can ensure the validity of InfoNCE loss with less memory space.
%\subsubsection{Momentum Based Update of Encoder}
\par Optimization of the auxiliary encoder is also a crucial problem, which profoundly affects the performance. It should have been included in the backpropagation, but due to the addition of memory bank, the optimization method has also been adjusted accordingly. In an extreme case, there can be two completely independent encoders, i.e., the main and auxiliary encoders with different parameters, and the error of InfoNCE loss should be used as their update rewards. Conversely, they can also be exactly the same, i.e., two encoders share the parameters. However, both of these situations will result in a rapidly changing auxiliary encoder. The role of the auxiliary encoder in PCLNet is to provide negative samples for the memory bank so as to support the calculation of InfoNCE loss. The negative sample partly plays the role of supervision information in the learning process, so the ones stored in the memory bank should have good consistency. In other words, for the stability of the training process, there should not be too much difference between the oldest and the latest individuals in the memory bank \cite{InstanceDiscrimination}. Therefore, a smooth changing auxiliary encoder is needed. Inspired by \cite{moco}, a momentum based method is employed to update the parameters of auxiliary encoder:
\begin{equation}
\left\{
\begin{aligned}
\tilde{\theta} &\leftarrow m \tilde{\theta}+(1-m) \theta\\
\tilde{\varphi} &\leftarrow m \tilde{\varphi}+(1-m) \varphi \label{5}
\end{aligned}
\right.
\end{equation}
where $m \in(0,1)$ means the momentum coefficient. In this way, a relatively large momentum encourages the auxiliary encoder to update more smoothly and stably. 
\par In summary, the InfoNCE loss of one mini-batch in PCLNet can be rewritten as:
\begin{equation}
\mathcal{L}(o,o^{+})\!=\!-\!\sum_{i=1}^{N} \!\log\! \frac{\exp \left(\left\langle{o_i, o_i^{+}} \right\rangle\! /\! \tau\right)}{\exp \left(\left\langle{o_i, o_i^{+}} \right\rangle \!/\! \tau\right)\!+\!\sum_{j=1}^{K} \exp \left(\left\langle{o_i, o_j^{-}} \right\rangle\! / \!\tau\!\right)}\label{4}
\end{equation}
where $o_i$ and $o_i^{+}$ are variables, and $o_j^{-}$ can be seen as a constant. Back propagating the above loss function via mini-batch SGD, the main encoder can be updated. Based on the evolved parameters of the main encoder, the auxiliary encoder can be updated through a momentum based method.
\subsection{Classifier Fine-tuning}
\begin{figure}[h]
	\begin{centering}
		\includegraphics[width=9.0cm, height=5.0cm]{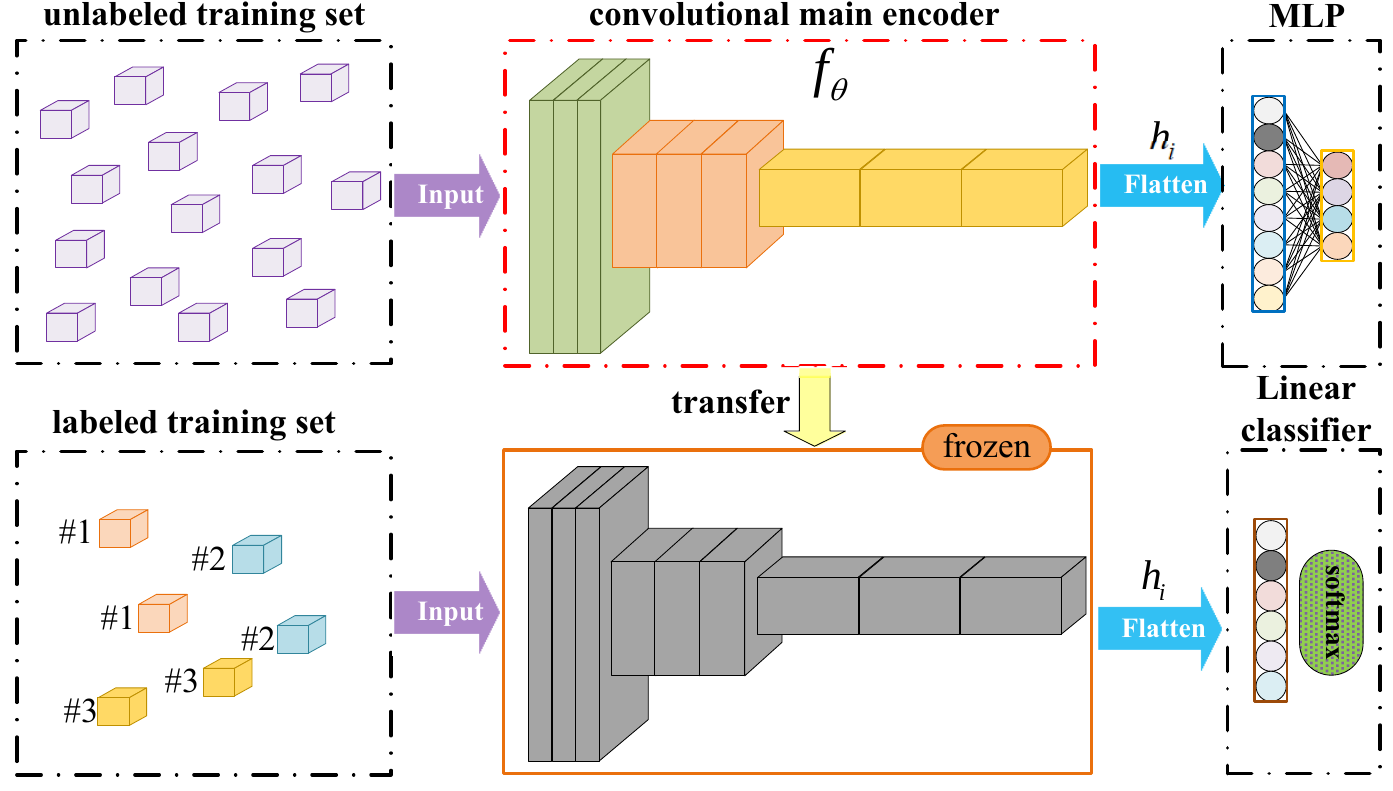}
		\caption{Illustration of classifier fine-tuning. The results of unsupervised pre-training are selectively used in the downstream task. Limited supervision is fed to train a new linear classifier, so as to achieve few-shot PolSAR classification.}\label{fig:g6}
	\end{centering}
\end{figure}
\par As known that completely relying on unsupervised learning may be difficult to meet the requirements of accuracy and flexibility. Because the obtained deep PolSAR representations cannot be directly used for classification. Therefore, fine-tuning a classifier with small amount of supervision is more acceptable for better performance when facing specific tasks. In this paper, fine-tuning generally follows the paradigm of supervised learning \cite{7762055}, but slightly different. The following of supervised learning is reflected in the dataset collection, the loss function definition and optimization method. The difference is that only the classifier is included in the training process of this part, and the feature extraction layers are not involved. So the number of training samples used for fine-tuning can be very small. The reason for this difference is that the representations learnt in unsupervised pre-training are transferrable, so that the dependence on complex paradigms and massive human annotations can be alleviated.
\par An illustration of the classifier fine-tuning of PCLNet is shown in Fig. \ref{fig:g6}. It can be seen that the pre-trained $f_{\theta}$ of main encoder is undoubtedly the foundation of few-shot classification. In the process of fine-tuning, $f_{\theta}$ will be used without any changes for the representation learning of labeled training samples. And a trainable linear classifier with a fully-connected layer followed by softmax activation is connected behind $f_{\theta}$. Limited supervision is sufficient to the training of linear classifier due to its low complexity. In summary, the whole training process of PCLNet is shown in Algorithm \ref{alg:2}. 
\begin{algorithm}[h!]
	\caption{Training Process of PCLNet}
	\label{alg:2}
	\begin{algorithmic}[1]
		\STATE \textbf{Begin}
		\STATE Prepare the contrastive learning dataset $\Theta=\{\mathbf{x}_i\}_{i=1}^{M}$. \\
		\REQUIRE Unsupervised pre-training\\
		\STATE Prepare the positive sample pair $\mathbf{x}_i$ and $\mathbf{x}_i^{+}$, max epoch $\mathcal{I}_1$, number of steps $B_1$, momentum coefficient $m$ and temperature $\tau$. Initialize learnable parameters $\theta\! =\!\tilde{\theta}$, $\varphi=\tilde{\varphi}$, and memory bank $\mathcal{Q}$. \\
		\FOR{iteration in $\mathcal{I}_1$:}
		\FOR{mini-batch in $B_1$:}
		\STATE $h_i=f_{\theta}\left(\mathbf{x}_i\right)$, $o_{i}=g_{\varphi}\left(h_{i}\right)$\\
		\STATE $\tilde{h}_i=f_{\tilde{\theta}}\left(\mathbf{x}_i^{+}\right)$, $o_{i}^{+}=g_{\tilde{\varphi}}\left(\tilde{h}_{i}\right)$\\
		\FOR{encoded negative sample $o_{j}^{-}$ in $\mathcal{Q}$:}
		\STATE Calculate the InfoNCE loss by \eqref{4}.\\
		\ENDFOR
		\STATE Optimize the parameters $\theta$ and $\varphi$ by mini-batch SGD of InfoNCE loss.
		\STATE Momentum update the parameters $\tilde{\theta}$ and $\tilde{\varphi}$ by \eqref{5}.
		\STATE Enqueue the current mini-batch $o^{-}$ and remove the oldest one to update $\mathcal{Q}$.
		\ENDFOR
		\ENDFOR
		\RETURN optimal parameter $\theta^{*}$ of $f_{\theta}$.
		\ENSURE Classifier fine-tuning\\
		\STATE Prepare the labeled PolSAR training set $\Omega=\{\left( \mathbf{I}_i, y_i\right) \}_{i=1}^{l}$, max epoch $\mathcal{I}_2$ and number of steps $B_2$. Freeze the optimal parameter $\theta^{*}$. Initialize learnable parameters $w$ and $b$ of a linear classifier.\\
		\FOR{iteration in $\mathcal{I}_2$:}		
		\FOR{mini-batch in $B_2$:}
		\STATE Optimize $w$ and $b$ with training set $\Omega$ by backpropagation.\\
		\ENDFOR
		\ENDFOR
		\RETURN optimal parameters $\theta^{*}$, $w^{*}$ and $b^{*}$.
	\end{algorithmic}
\end{algorithm}
\section{Experimental Results}\label{sec:exp}
\subsection{Datasets Description}
\begin{figure}[h]
	\begin{centering}
		\includegraphics[width=0.48\textwidth]{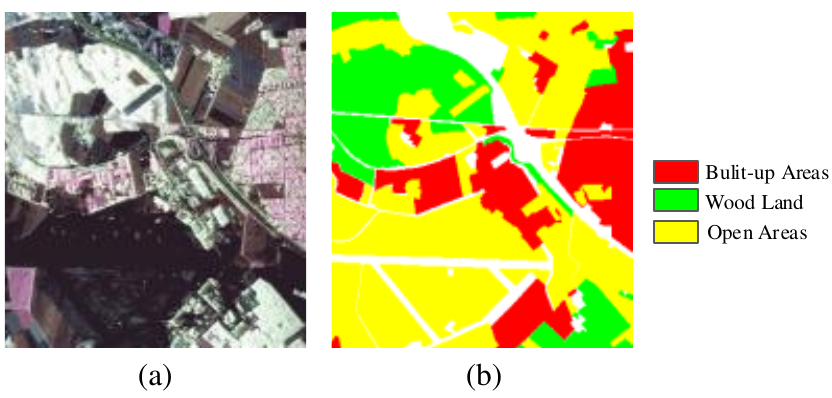}
		\caption{ESAR Oberpfaffenhofen dataset and its color code. (a) Pauli RGB map. (b) Ground truth map.}\label{fig:gt2}
	\end{centering}
\end{figure}

We employ two widely-used PolSAR datasets in the experiments, i.e., AIRSAR Flevoland and ESAR Oberpfaffenhofen. Figs. \ref{fig:gt1}-\ref{fig:gt2} show their Pauli and ground truth maps respectively. Besides, Tables \ref{tab:tab1}-\ref{tab:tab2} show some details about the benchmark datasets.
\subsubsection{AIRSAR Flevoland}
\begin{figure}[h]
	\begin{centering}
		\includegraphics[width=0.5\textwidth]{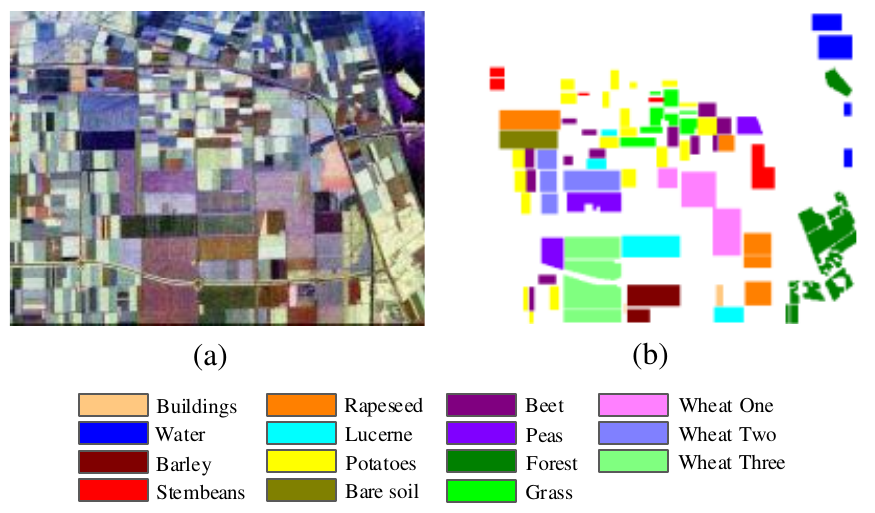}
		\caption{AIRSAR Flevoland dataset and its color code. (a) Pauli RGB map. (b) Ground truth map.}\label{fig:gt1}
	\end{centering}
\end{figure}

As shown in Fig. \ref{fig:gt1}, an L-band, full polarimetric image of the agricultural region of the Netherlands is obtained through NASA/Jet Propulsion Laboratory AIRSAR. The size of this image is $750\times1024$ and the spatial resolution is $0.6m\times1.6m$. There are $15$ kinds of ground objects including buildings, stembeans, rapeseed, beet, bare soil, forest, potatoes, peas, lucerne, barley, grasses, water and three kinds of wheat. The number of the labeled pixels can be seen in Table \ref{tab:tab1}.
\begin{table}[h]
	\centering  
	\caption{Number of Pixels in Each Category for Airsar Flevoland}
	\renewcommand\arraystretch{1.25}
	\begin{tabular}{p{54pt}|p{54pt}|p{54pt}c} 
		\Xhline{2pt}
		\multicolumn{3}{c}{\centering AIRSAR Flevoland} \\
		\hline
		\centering Class code &Name &Reference data\\ \hline 
		\centering 1 &Buildings  &963 \\        
		\centering 2 &Rapeseed  &17195  \\       
		\centering 3 &Beet &11516\\
		\centering 4 &Stembeans &6812  \\
		\centering 5 &Peas &11394  \\
		\centering 6 &Forest &20458 \\
		\centering 7 &Lucerne &11411  \\
		\centering 8 &Potatoes &19480  \\
		\centering 9 &Bare soil &6116  \\
		\centering 10 &Grass &8159  \\
		\centering 11 &Barley &8046  \\
		\centering 12 &Water &8824  \\
		\centering 13 &Wheat one &16906  \\
		\centering 14 &Wheat two &12728  \\
		\centering 15 &Wheat three &24584  \\ \hline
		\centering Total & - &184592  \\
		\Xhline{2pt}
	\end{tabular}
	\label{tab:tab1}
\end{table}
\subsubsection{ESAR Oberpfaffenhofen}
An L-band, full polarimetric image of Oberpfaffenhofen, Germany, $1200\times 1300$ scene size, are obtained through ESAR airborne platform. Its Pauli color-coded image and ground truth map can be seen in Fig. \ref{fig:gt2}. There are three kinds of ground objects in the ground truth map including built-up areas, wood land and open areas. The number of labeled pixels can be seen in Table \ref{tab:tab2}.
\begin{table}[!t]
	\centering  
	\caption{Number of Pixels in Each Category for Esar Oberpfaffenhofen}
	\renewcommand\arraystretch{1.25}
	\begin{tabular}{p{54pt}|p{54pt}|p{54pt}c} 
		\Xhline{2pt}
		\multicolumn{3}{c}{\centering ESAR Oberpfaffenhofen} \\
		\hline
		\centering Class code &Name &Reference data\\ \hline 
		\centering 1 &Built-up areas  &310829 \\        
		\centering 2 &Wood land  &263238  \\
		\centering 3 &Open areas  &733075  \\ \hline
		\centering  Total&-  &1307142  \\
		\Xhline{2pt}
	\end{tabular}
	\label{tab:tab2}
\end{table}
\subsection{Experimental Setup}
\subsubsection{Data Preparations}
The original PolSAR images are represented by the polarimetric coherency matrix $\mathbf{T}$. In the diversity stimulation mechanism, the cluster numbers of AIRSAR and E-SAR datasets are set to 35 and 50, respectively. Then, affinity based on the Gaussian kernel function is used to express the similarity between instances, and the value of bandwidth is set to 0.42. Finally, 600 instances are filtered out from each cluster as the training set. In the pretext task, the upper triangular elements of $\mathbf{T}$ are divided into the real and imaginary parts to describe each pixel of the PolSAR image. In the fine-tuning stage, the process is similar to some traditional methods. Not only the pixels, but also the surrounding $15 \times 15$ image patches are cropped to generate the datasets. Then, the training sets with 300 samples, validation sets with 200 samples and testing datasets with the remaining samples are obtained.  
\subsubsection{Parameter Settings and Comparing Methods}
\begin{table*}[!t]
	\centering  
	\caption{Comparisons of Full-supervised Classification Results ($\%$) for Airsar Flevoland Dataset.}
	\renewcommand\arraystretch{1.25}
	\begin{tabular}{C{2.0cm}C{1.1cm}C{1.1cm}C{1.1cm}C{1.1cm}C{1.1cm}C{1.1cm}C{1.1cm}C{1.1cm}C{1.1cm}} 
		\Xhline{2pt}
		Method&Wishart&SVM&MLP&CNN&CV-CNN&SF-CNN&TFL&MAML&PCLNet\\
		\midrule									
		Buildings&91.59&98.75&88.27&92.83&97.72&95.02&94.18&94.18&95.43\\
		Rapeseed&73.70&73.72&83.12&70.89&64.90&84.48&59.16&81.00&87.11\\
		Beet&93.67&86.28&84.43&88.00&89.12&67.83&71.28&97.35&96.41\\
		Stembeans&89.68&92.84&93.94&94.38&98.08&98.80&96.21&89.74&96.59\\
		Peas&91.45&83.31&83.02&97.14&95.73&94.33&94.28&82.30&97.96\\
		Forest&85.51&79.44&89.26&97.67&98.09&97.44&87.69&96.82&98.15\\
		Lucerne&78.11&86.55&85.97&81.78&90.75&97.26&97.37&84.48&95.22\\
		Potatoes&92.74&79.30&83.40&96.67&89.19&89.53&93.13&86.96&96.00\\
		Bare soil&70.85&95.24&93.84&75.20&92.81&99.71&61.59&92.02&92.81\\
		Grass&24.86&61.10&60.46&83.52&94.72&53.27&17.17&94.25&85.98\\
		Barley&99.14&87.81&96.07&82.12&70.40&95.20&69.94&84.41&93.84\\
		Water&57.54&98.70&98.16&99.34&76.30&99.98&98.54&95.58&95.26\\
		Wheat one&94.64&72.76&82.08&91.57&93.67&98.86&84.58&89.95&94.33\\
		Wheat two&36.11&73.44&74.96&82.16&92.11&80.68&58.96&82.10&84.10\\
		Wheat three&81.86&69.86&80.24&85.06&96.40&92.24&86.15&83.32&96.60\\
		\midrule									
		OA&78.81&79.29&84.10&88.03&89.28&89.81&79.23&88.09&\textbf{93.96}\\
		AA&77.43&82.61&85.15&87.89&89.33&89.64&78.02&88.96&\textbf{93.72}\\
		Kappa&77.53&78.17&83.11&87.19&88.50&89.06&77.96&87.27&\textbf{93.47}\\
		\Xhline{2pt}
	\end{tabular}
	\label{tab:tab4}
\end{table*}
At the beginning of pre-training, each training sample is considered as $x_i$, and the corresponding positive sample $\tilde{x}_i$ is generated by rotating $180^{\circ}$. The parameter settings of main encoders and auxiliary encoders are identical, which is crucial for retaining consistency. The detailed information is displayed in Fig. \ref{fig:g4}. For the convolutional encoder, the size of the convolution kernels is $3\times3$ with stride 1. And the number of the kernels in three convolution layers is 16, 32 and 64, respectively. The size of max pooling is $2\times2$ and the stride is 2. For the projection heads, the dimensions of two fully-connected layers are 64 and 32, which means the extracted feature vectors are 32-dimensional here. SGD is employed in our experiments to optimize the models. We implement pre-training for 800 epochs with the initial learning rate of 0.1, and the learning rate will be multiplied by 0.5 at the 300 and 500 epochs. At the same time, the mini-batch size is set as 512 while the length of memory bank is 8192. Besides, the momentum coefficient $m$ is 0.999 and the temperature $\tau$ is set to be 0.4.
\par When the network pre-training is finished, a linear classifier, i.e., a fully connected to softmax layer, is added behind the GAP layer, and its dimension is equal to the number of categories. The classifier is trained for 300 epochs with the learning rate of 0.01 and the mini-batch size of 32. Our code is available at: https://github.com/Siyuzhang-hit/PCLNet.
\par In order to evaluate the effectiveness of the proposed method, several supervised and semi-supervised classifiers are performed and tested in the experiments. Specifically, three classical shallow models with hand-crafted features are chosen, including Wishart \cite{super-wishart}, radial basis function kernel based support vector machine (SVM) \cite{svm}. Four CNN-based methods including MLP \cite{5653650}, CNN \cite{7762055}, CV-CNN \cite{complex} and polarimetric-feature-driven CNN (SF-CNN) \cite{input1} are chosen for comparison. Moreover, two representative few-shot learning methods, i.e., transfer learning and meta learning, are selected to be compared. In this paper, transfer learning is realized by a ImageNet pre-trained VGG-11 architecture \cite{vgg}, and meta learning is realized by model-agnostic meta-learning \cite{maml}. We denote them as TFL and MAML for convenience.
\iffalse
\subsubsection{Evaluation Criteria}
To evaluate the classification performance quantitatively, we chose three standard criteria including overall accuracy (OA), average accuracy (AA) and kappa coefficient (Kappa). They can be defined as follows:
\begin{equation}
\text{OA}=\frac{\sum_{i=1}^c M_i}{\sum_{i=1}^c N_i}, \; \text{AA}=\frac{1}{c}\sum_{i=1}^c\frac{M_i}{N_i} 
\label{eqexp:3}
\end{equation}
where $c$ denotes category numbers. $M_i$ means the number of correctly classified samples of the $i$th category and $N_i$ denotes the number of the $i$th labeled samples.
\begin{equation}
\text{Kappa}=\frac{\text{OA}-P}{1-P}, \; \text{with} \; P=\frac{1}{N^2}\sum_{i=1}^cH(i,:)H(:,i)
\label{eqexp:4}
\end{equation}
where $N$ is the number of testing samples and $H$ denotes the classification confusion matrix.
\fi
\subsection{Experimental Results}
\subsubsection{Classification Results}
\begin{figure*}[!t]
	\begin{centering}
		\includegraphics[width=\textwidth]{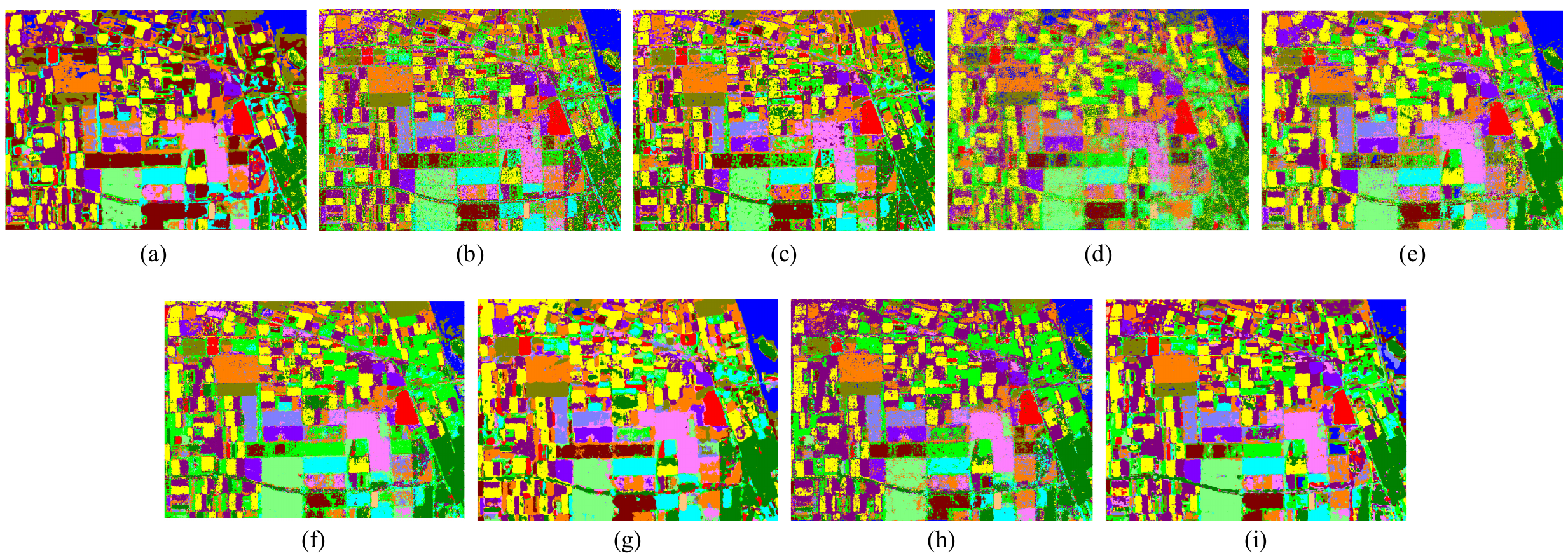} %,height=8.0cm
		\caption{Full-supervised classification results of the whole map on AIRSAR Flevoland dataset with different methods. (a) Result of Wishart. (b) Result of SVM. (c) Result of MLP. (d) Result of CNN. (e) Result of CV-CNN. (f) Result of SF-CNN. (g) Result of TFL. (h) Result of MAML. (i) Result of PCLNet.}\label{fig:3}
	\end{centering}
\end{figure*}
\begin{figure*}[!t]
	\begin{centering}
		\includegraphics[width=\textwidth]{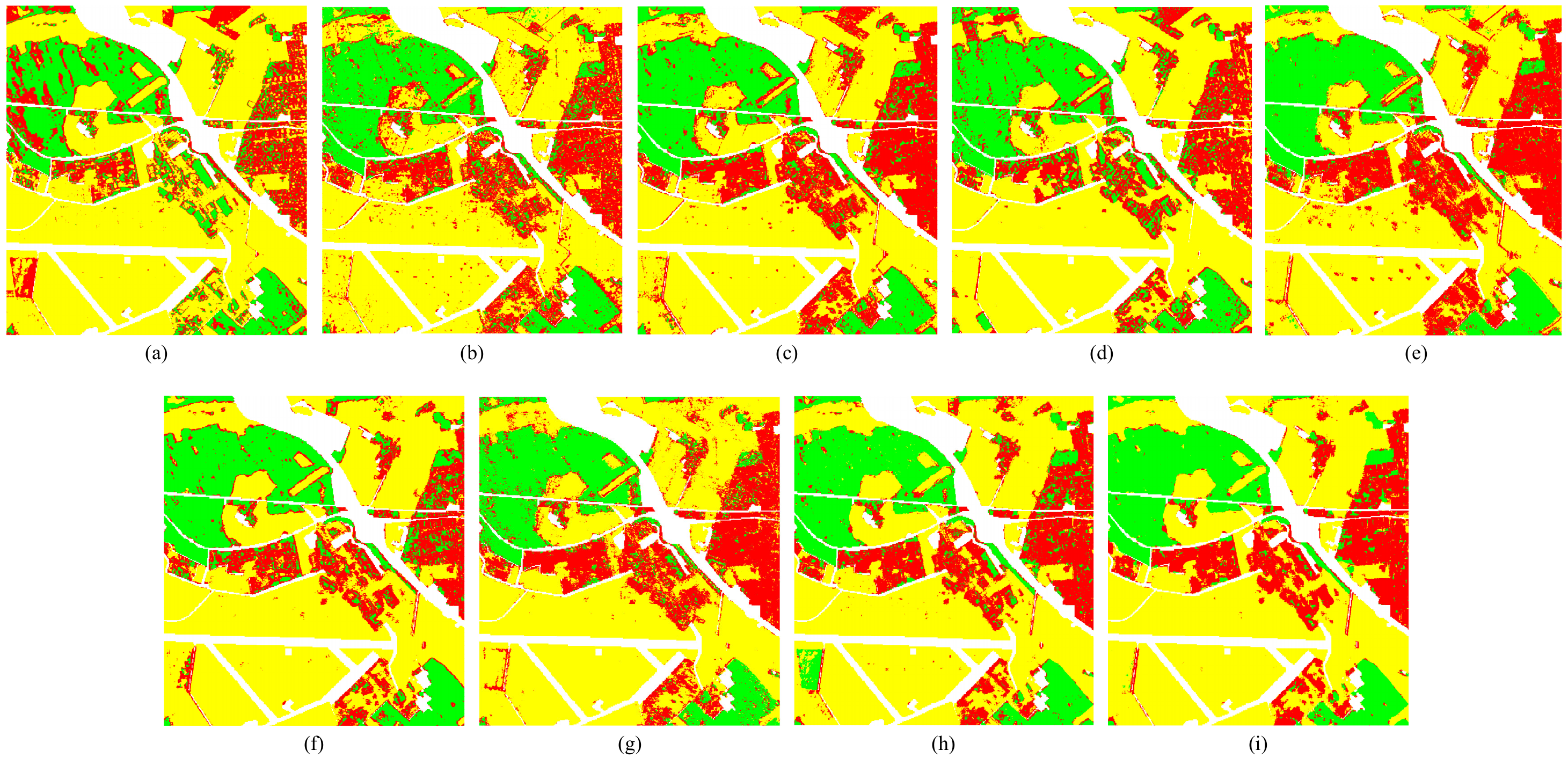} %
		\caption{Full-supervised classification results of the whole map on ESAR Oberpfaffenhofen dataset with different methods. (a) Result of Wishart. (b) Result of SVM. (c) Result of MLP. (d) Result of CNN. (e) Result of CV-CNN. (f) Result of SF-CNN. (g) Result of TFL. (h) Result of MAML. (i) Result of PCLNett.}\label{fig:4}
	\end{centering}
\end{figure*}

\begin{table*}[!t]
	\centering  
	\caption{Comparisons of Few-shot Classification Results ($\%$) on Airsar Flevoland Dataset.}
	\renewcommand\arraystretch{1.25}
	%\begin{tabular}{C{1.4cm}C{1.0cm}C{1.0cm}C{1.3cm}C{1.0cm}C{1.0cm}C{1.1cm}C{1.1cm}C{1.0cm}C{1.0cm}C{1.0cm}} 
	\begin{tabular}{C{2.0cm}C{1.1cm}C{1.1cm}C{1.1cm}C{1.1cm}C{1.1cm}C{1.1cm}C{1.1cm}C{1.1cm}C{1.1cm}} %11.6
		\Xhline{2pt}
		Method&Wishart&SVM&MLP&CNN&CV-CNN&SF-CNN&TFL&MAML&PCLNet\\
		\midrule									
		Buildings&69.26&94.29&79.44&89.20&79.02&87.75&89.72&97.72&94.39 \\
		Rapeseed&25.12&71.24&75.44&34.78&51.78&73.71&55.76&81.82&69.28 \\
		Beet&52.33&81.86&90.04&43.74&53.27&75.94&62.71&92.94&91.76 \\
		Stembeans&60.47&83.43&87.40&75.25&70.96&95.41&86.10&97.03&92.35 \\
		Peas&37.37&78.00&91.81&70.58&56.75&87.52&69.14&85.97&93.81 \\
		Forest&48.77&72.31&91.49&57.71&60.06&90.04&68.61&82.66&96.44 \\
		Lucerne&43.65&83.05&89.15&84.87&66.01&97.91&62.31&90.82&94.00 \\
		Potatoes&42.18&74.67&66.14&47.46&42.70&70.45&65.64&84.44&91.76 \\
		Bare soil&75.43&93.18&87.34&69.00&74.92&92.04&74.26&93.80&85.79 \\
		Grass&27.52&41.93&52.57&42.81&43.03&60.42&55.57&60.72&64.43 \\
		Barley&66.31&70.90&83.97&79.01&84.19&84.97&75.25&90.17&94.93 \\
		Water&58.39&97.05&76.56&77.69&72.98&76.87&85.51&99.29&85.97 \\
		Wheat one&41.19&64.02&47.37&62.84&71.53&54.54&65.11&68.25&89.07 \\
		Wheat two&35.69&63.98&65.26&52.33&36.18&28.06&56.25&68.75&83.09 \\
		Wheat three&22.69&55.74&35.52&59.37&66.28&67.08&63.41&87.95&90.12 \\
		\midrule									
		OA&41.71&71.52&70.69&58.82&59.34&70.76&65.97&83.68&\textbf{87.88} \\
		AA&47.09&75.04&74.63&63.11&61.98&72.18&69.02&85.49&\textbf{87.81} \\
		Kappa&40.61&70.25&69.40&57.45&57.91&69.35&64.61&82.65&\textbf{87.02} \\
		\Xhline{2pt}
	\end{tabular}
	\label{tab:tab3}
\end{table*}
In the experiments, 20 and 300 training samples of each category are used to perform the few-shot and full-supervised PolSAR image classification, respectively. To evaluate the classification performance quantitatively, three criteria including overall accuracy (OA), average accuracy (AA) and kappa coefficient (Kappa) are chosen. Tables \ref{tab:tab4}-\ref{tab:tab3} report the classification results on Flevoland dataset with the aforementioned experimental settings, and Tables \ref{tab:tab6}-\ref{tab:tab5} for Oberpfaffenhofen dataset. Moreover, the classification maps are shown in Figs. \ref{fig:3} and \ref{fig:4}. Generally speaking, different situations have different results, but the trends of different datasets are consistent. Hence, we describe the results from two aspects. In the case of few-shot classification, the performance of traditional Wishart classifier is not satisfactory. MLP and SVM show better results than CNN and CV-CNN. The performance of SF-CNN and MAML is slightly improved compared with the traditional CNN-based methods. In addition, the knowledge mining from optical images by TFL is not suitable to transfer into PolSAR images. As the number of training samples increases, the performance of CNN-based methods is improved and surpasses other methods such as Wishart, SVM, MLP and MAML. The results of TFL are still not satisfactory. The proposed PCLNet emerges the best generalization performance in two cases. The specific analysis of the experimental results and classification maps on each dataset is as follows.
\par Full-supervised classification results of the whole map on Flevoland dataset are presented in Fig. \ref{fig:3}. As mentioned above, totally 300 (about $2.44\%$ sampling rate) labeled samples of each category are utilized as the supervision to fine-tune the classifier. It can be observed from these results that the proposed PCLNet achieves the best completeness of the terrains. Moreover, the situations where buildings are incorrectly assigned to forest or intersection locations of different terrains can be reduced, which confirms the validity of the proposed PCLNet. 
\begin{figure}[!t]
	\begin{centering}
		\includegraphics[width=0.49\textwidth]{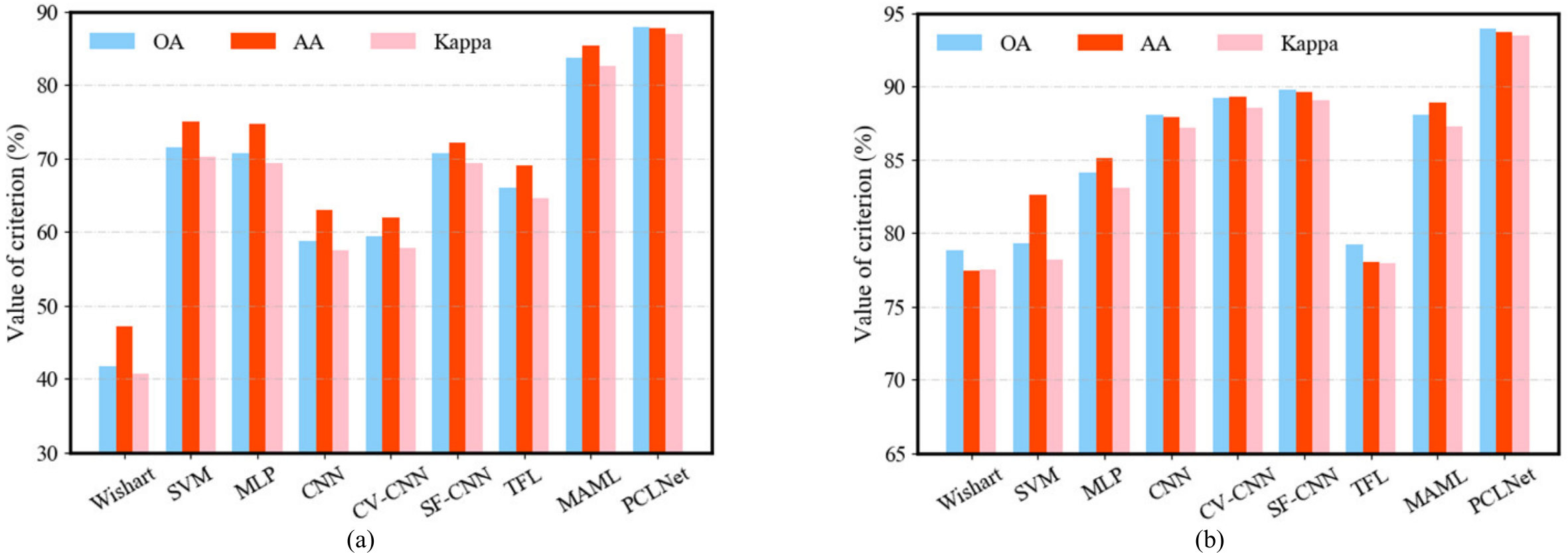} %
		\caption{Comparisons of involved methods on AIRSAR Flevoland dataset. (a) Result of few-shot classification. (b) Result of full-supervised classification.}\label{fig:1}
	\end{centering}
\end{figure} 
\begin{table*}[!t]
	\centering  
	\caption{Comparisons of Full-supervised Classification Results ($\%$) for Esar Oberpfaffenhofen Dataset.}
	\renewcommand\arraystretch{1.25}
	\begin{tabular}{C{2.0cm}C{1.1cm}C{1.1cm}C{1.1cm}C{1.1cm}C{1.1cm}C{1.1cm}C{1.1cm}C{1.1cm}C{1.1cm}}
		\Xhline{2pt}
		Method&Wishart&SVM&MLP&CNN&CV-CNN&SF-CNN&TFL&MAML&PCLNet\\
		\midrule									
		Built-up areas&51.21&73.32&80.25&84.52&85.81&79.38&65.22&74.63&86.84 \\
		Wood land&72.20&88.90&91.27&92.94&93.34&86.20&86.08&93.94&95.38 \\
		Open areas&94.86&88.13&89.29&89.39&88.75&92.64&91.07&92.39&91.19 \\
		\midrule									
		OA&79.92&84.76&87.54&88.95&88.97&88.19&83.92&89.03&\textbf{92.50} \\
		AA&72.76&83.45&86.94&88.95&89.30&86.07&80.79&86.99&\textbf{91.13} \\
		Kappa&70.13&77.87&73.74&76.72&81.80&82.14&76.34&83.23&\textbf{87.36} \\
		\Xhline{2pt}
	\end{tabular}
	\label{tab:tab6}
\end{table*}
\par Quantitative comparisons are reported in Tables \ref{tab:tab4}-\ref{tab:tab3}, in which the proposed method achieves the highest scores on three criteria. PCLNet improves OA, AA, and Kappa of CNN by $5.93\%$, $5.83\%$, and $6.28$; $4.68\%$, $4.39\%$, and $4.97$ increase of OA, AA, and Kappa are accomplished for complex-valued CNN. Furthermore, the results with only 20 (about $0.16\%$ sampling rate) training samples for each category also demonstrates the effectiveness of PCLNet. As shown in Table \ref{tab:tab3}, the best results obtained by the proposal can reach $87.88\%$ OA, $87.81\%$ AA, and $87.02$ Kappa, and these scores are almost equivalent to using CNN for full-supervised classification. Combined with the results of the two cases, it is not difficult to find that the performance of traditional CNN-based methods is seriously limited by the number of labeled training samples. Accompanied by the decrease in the number of training samples, only $6.08\%$ OA, $5.91\%$ AA, and $6.45$ Kappa of PCLNet are reduced, but the values of CNN dropped by $29.21\%$, $24.78\%$, and $29.74$. Moreover, the accuracy of most categories decreases dramatically not only in traditional CNN-based methods, but also in shallow models. However, the proposed method can better maintain the classification performance in the case of insufficient supervision information.
\begin{table*}[!t]
	\centering  
	\caption{Comparisons of Few-shot Classification Results ($\%$) on Esar Oberpfaffenhofen Dataset.}
	\renewcommand\arraystretch{1.25}
	\begin{tabular}{C{2.0cm}C{1.1cm}C{1.1cm}C{1.1cm}C{1.1cm}C{1.1cm}C{1.1cm}C{1.1cm}C{1.1cm}C{1.1cm}} %12.3
		\Xhline{2pt}
		Method&Wishart&SVM&MLP&CNN&CV-CNN&SF-CNN&TFL&MAML&PCLNet\\
		\midrule									
		Built-up areas&46.62&61.89&62.16&50.55&51.44&70.57&59.35&73.84&78.06 \\
		Wood land&70.82&85.85&88.33&90.00&90.85&71.73&69.94&82.63&90.72 \\
		Open areas&92.99&85.71&89.05&91.67&92.62&89.94&92.24&90.73&84.92 \\
		\midrule									
		OA&77.50&80.07&82.51&81.56&81.40&81.67&79.93&85.08&\textbf{86.54} \\
		AA&70.14&77.82&79.85&77.41&78.30&77.41&73.84&82.40&\textbf{84.57} \\
		Kappa&67.06&71.92&63.14&61.13&69.41&73.41&70.70&76.00&\textbf{77.65} \\
		\Xhline{2pt}
	\end{tabular}
	\label{tab:tab5}
\end{table*}
\begin{figure}[!t]
	\begin{centering}
		\includegraphics[width=0.49\textwidth]{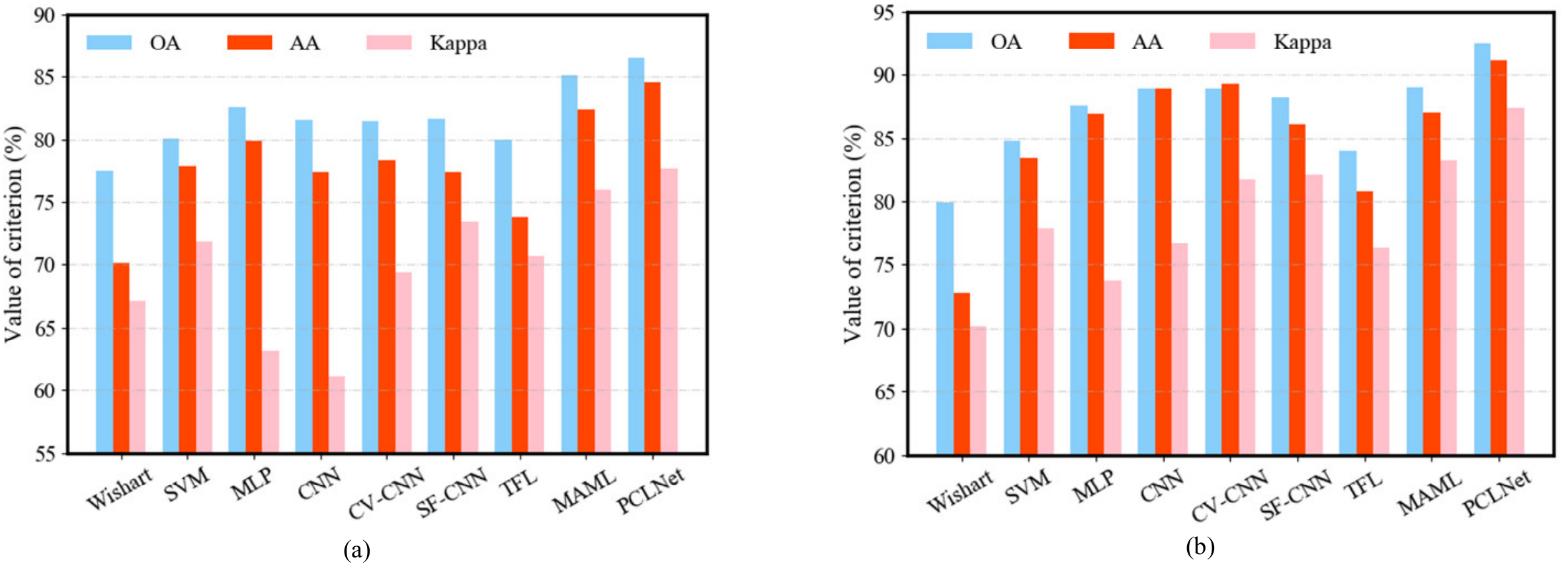} %
		\caption{Comparisons of involved methods on ESAR Oberpfaffenhofen dataset. (a) Result of few-shot classification. (b) Result of full-supervised classification.}\label{fig:2}
	\end{centering}
\end{figure}
\par In order to compare more clearly, Fig. \ref{fig:1} shows the performance comparisons of few-shot classification and full-supervised classification on AIRSAR Flevoland dataset. These results reveal that high-level representations learnt by CL can effectively alleviate the greedy demands of CNN-based methods for massive annotations. At the same time, the relatively low complexity of the linear classifier fine-tuned in the downstream task can better avoid the occurrence of overfitting. To sum up, the experimental results on Flevoland dataset can confirm the effectiveness of PCLNet.
\par Fig. \ref{fig:4} displays the full-supervised classification results of the whole map on ESAR Oberpfaffenhofen dataset. The same as before, 300 labeled samples of each category (sampling rate is about 0.69\textperthousand) are used in the classifier fine-tuning stage. As shown in the classification maps, compared with other classifiers, the proposal can better distinguish the built-up areas and wood land, and fewer misclassification points are contained in the open areas. Other methods depict more errors especially in the wood land. Tables \ref{tab:tab6}-\ref{tab:tab5} summarize the experimental results of each method quantitatively, and Fig. \ref{fig:2} shows the performance comparisons of different methods on ESAR Oberpfaffenhofen dataset. In the case of full-supervised classification, the proposed PCLNet achieves the increments of $3.55\%$ OA, $2.18\%$ AA, $10.64$ Kappa for CNN, and $3.53\%$ OA, $1.83\%$ AA, $5.56$ Kappa for complex-valued CNN. And higher scores are obtained by the proposal than those from other classical and CNN-based algorithms. In the case of few-shot classification, only 20 labeled samples for each category (sampling rate is about 0.05\textperthousand) are used to when constructing the training sets. The accuracy on testing sets can achieve $86.54\%$ OA, $84.57\%$ AA, and $77.65$ Kappa for PCLNet. As shown in Fig. \ref{fig:2}, the performance of PCLNet on few-shot learning is close to that of traditional CNN-based classifiers under full supervision. It can be seen that the smaller the number of training samples, the more obvious the advantages of PCLNet. Therefore, the experimental results can verify the validity of the proposal to a certain extent.
\begin{figure*}[!t]
	\begin{centering}
		\includegraphics[width=\textwidth]{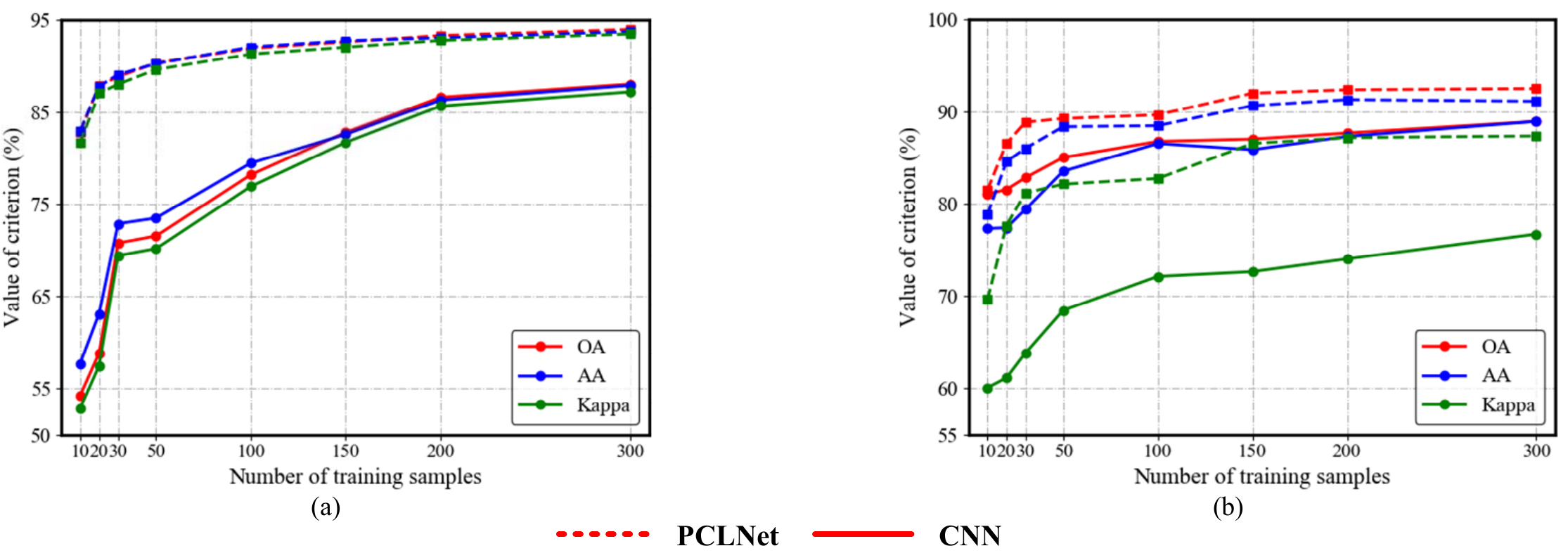} %
		\caption{Comparisons of the performance with different numbers of training samples between CNN and PCLNet on two benchmark datasets. The solid and dotted lines represent the results of CNN and PCLNet, respectively. (a) Result of Flevoland dataset. (b) Result of Oberpfaffenhofen dataset.}\label{fig:7}
	\end{centering}
\end{figure*}
\begin{figure*}[!t]
	\begin{centering}
		\includegraphics[width=\textwidth]{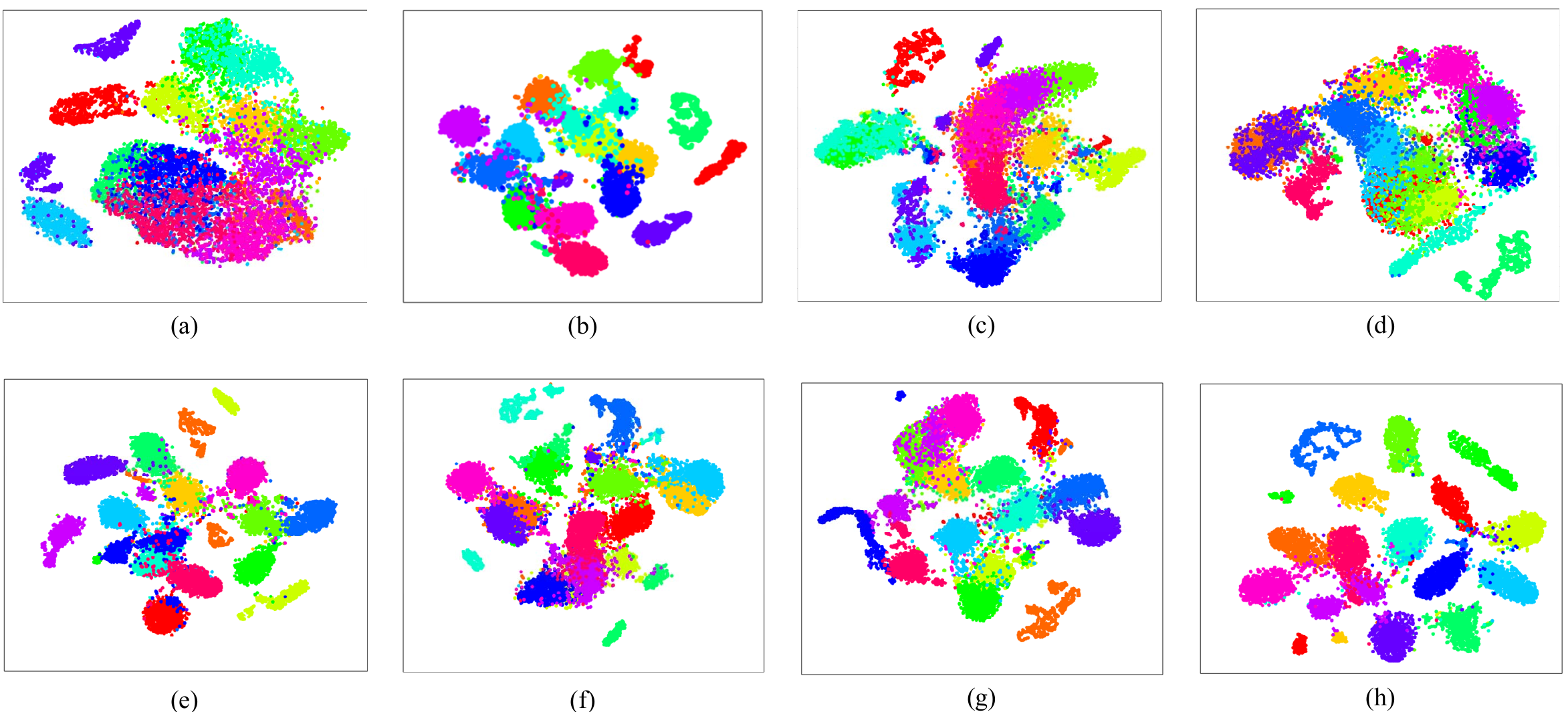} %,height=8.0cm
		\caption{T-SNE visualization of the representations learnt with different methods on AIRSAR Flevoland dataset. (a) Result of hand-crafted features. (b) Result of MLP. (c) Result of CNN. (d) Result of CV-CNN. (e) Result of SF-CNN. (f) Result of TFL. (g) Result of MAML. (h) Result of PCLNet. Each data point in the t-SNE scatter plots is colored according to its ground truth map.}\label{fig:5}
	\end{centering}
\end{figure*}
\subsubsection{Impact of Shots}
In order to investigate the influence of the number of training samples, comparative experiments are carried out. Specifically, CNN and the proposed PCLNet are tested under the environment of 10, 20, 30, 50, 100, 150, 200 and 300 training samples for each category. Fig. \ref{fig:7} reports the comparative results on two benchmark datasets. It can be seen from the results that the performance of PCLNet is better than that of CNN in all conditions. Especially on the Flevoland dataset, when only 10 samples for each category are used, the performance gap between two methods is the largest, which is $38.56\%$, $25.25\%$, and $29.1$ in terms of OA, AA, and Kappa. In contrast, if CNN wants to achieve the same classification performance, it needs at least 150 training samples for each category. On the Oberpfaffenhofen dataset, the significant improvement of Kappa coefficients can also demonstrate the validity of the proposed method in the case of few-shot learning.
\subsubsection{Visualization of Features}
\begin{figure*}[!t]
	\begin{centering}
		\includegraphics[width=\textwidth]{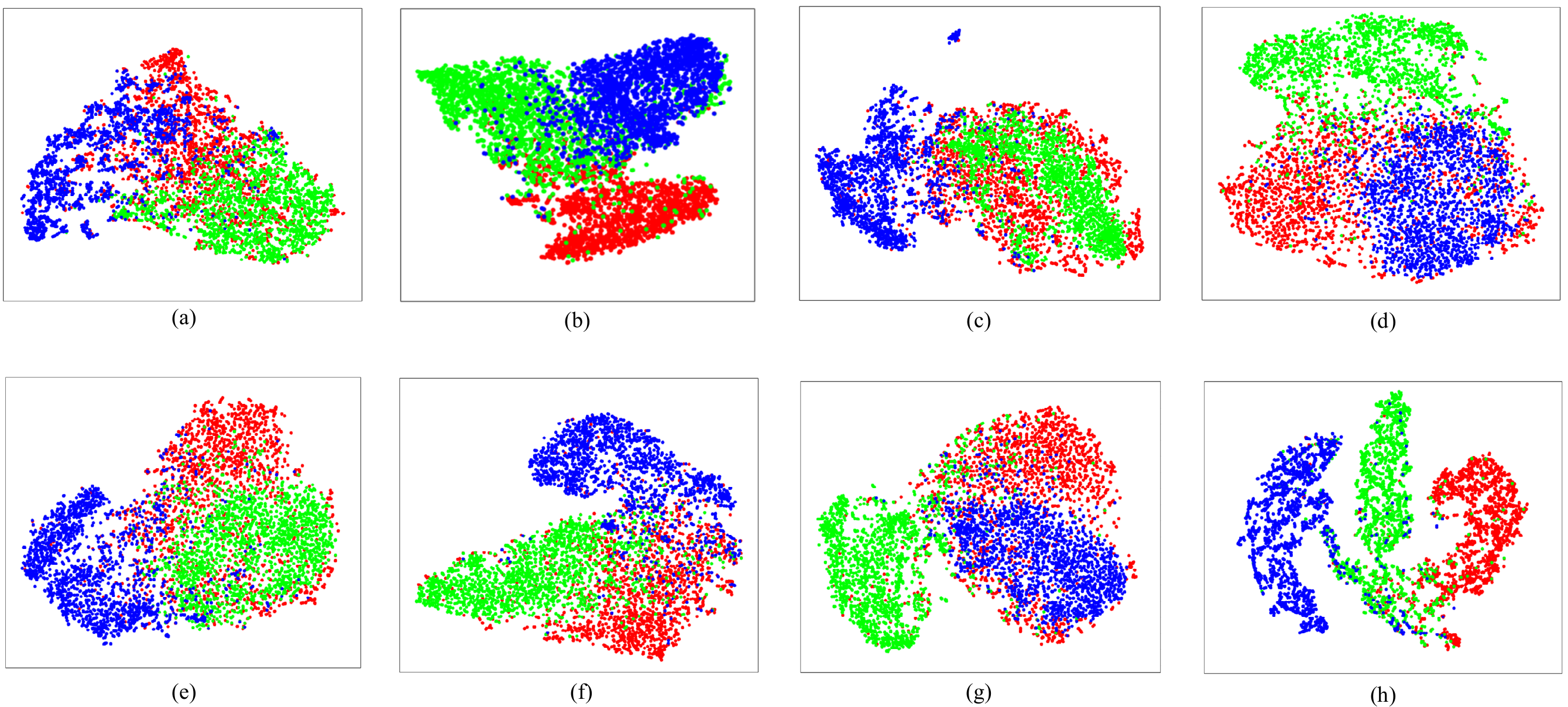} %,height=8.0cm
		\caption{T-SNE visualization of the representations learnt with different methods on ESAR Oberpfaffenhofen dataset. (a) Result of hand-crafted features. (b) Result of MLP. (c) Result of CNN. (d) Result of CV-CNN. (e) Result of SF-CNN. (f) Result of TFL. (g) Result of MAML. (h) Result of PCLNet. Each data point in the t-SNE scatter plots is colored according to its ground truth map.}\label{fig:6}
	\end{centering}
\end{figure*}

PCLNet has presented promising performance with limited labeled training samples, the intrinsic reason is the mining of high-level representations in the pretext task of instance discrimination as well as the utilization of unlabeled PolSAR images. In order to evaluate the quality of extracted features, two-dimensional visualizations of the learnt representations are performed by t-SNE \cite{tsne}. In this experiment, MLP, CNN, CV-CNN, SF-CNN, TFL and MAML utilize 20 samples for each category in the phase of training. We visualize the activation responses of the last hidden layer in these network models. Besides, hand-crafted features used for the training of Wishart and SVM are also visualized, including Freeman-Durden three-component decomposition \cite{freeman}, Yamaguchi four-component decomposition \cite{oldmethod3}, and H/A/$\alpha$ \cite{halpha}. For PCLNet, the deep PolSAR representations learnt through the unsupervised pre-training are visualized without the participation of any labeled samples. These experiments are implemented on two benchmark datasets separately, and the final t-SNE visualization results are shown in Figs. \ref{fig:5}-\ref{fig:6}. The color-coding of the scatter points is consistent with the ground truth maps.
\par It can be observed that the hand-crafted features are not adequate for distinguishing most categories, which produces great challenges to the design of appropriate classifiers. For the results of MLP, some features from the same category are extensively distributed in various positions and develop multiple disconnected regions, so the compactness is relatively weak. For CNN method, the compactness improves slightly, but numerous points from different categories may overlap and cover with each other seriously. Some advanced CNN-based methods improve the separability of each category. SF-CNN slightly raises the feature quality by mapping the hand-crafted features from the original space to a high-dimensional embedding space. TFL and MAML improve their performances through the knowledge transfer. Compared with other methods, PCLNet provides more discriminative features and creates more compact and distinctive category-specific clusters. It turns out that the proposal can embed original feature vectors into a more discriminative space, which provides support for better generalization ability.
\par The experimental results illustrated above exhibit the advantages of unsupervised deep PolSAR representation learning. Firstly, the pretext task of instance discrimination supports the feature extractors to capture more discriminative semantic cues. That is exactly what supervised learning hopes to get through human annotations. Secondly, InfoNCE loss based on cosine similarity successfully assists the encoder to obtain some feature vectors with low intra-class variance \cite{cosine}. So the transferrable representations corresponding to the same category compactly match with the weight vectors of that category, which brings great convenience to the classifier fine-tuning for downstream tasks. Finally, nonlinear projection head effectively avoids the loss of information induced by InfoNCE loss. Hence, more discriminative representations can be sufficiently produced and maintained.
\subsection{Discussion}
In the above experiments, the high-level transferrable representations captured by PCLNet present powerful generalization abilities. At the same time, the performance of the proposal has made a significant breakthrough in few-shot PolSAR classification. Therefore, it is necessary to combine the theoretical basis and experimental results to analyze and discuss the proposed method comprehensively.
\par First of all, a diversity stimulation mechanism is assembled to collect the datasets used for unsupervised pre-training. This component makes it possible to take full advantage of massive unlabeled PolSAR data and improve the correctness of negative sampling in instance discrimination. We maintain that the improvement of diversity is the key factor to unlock the  bottleneck of the application gap between optics and PolSAR. 
\par Secondly, high-level representations alleviate the greedy demands of CNNs for abundant human annotations. The t-SNE scatter plots show that the proposal creates more compact and separable clusters, indicating that the transferrable representations are learnt through discovering the distinction between individuals. Although no augmentation or regularization techniques are used, the proposed method can still achieve promising results in few-shot PolSAR classification. By contrast, training traditional CNN-based PolSAR classifiers from scratch with limited training samples is easy to overfit the training data and affect the generalization performance.
\par Last but not least, among all of the supporting evidence, unsupervised representation learning, which combines the discrimination ability of CNN-based methods and the feasibility of unsupervised methods, can undoubtedly be extended to more downstream tasks. In this paper, we sufficiently reveal the advantages of PCLNet in the high-precision few-shot PolSAR image classification. However, the capacity of unsupervised representation learning is not limited to a single specific task, but also lies in the broad application fields.
\section{Conclusion}\label{sec:con}
In this paper, a practical way for unsupervised PolSAR representation learning and few-shot classification is explored with the help of CL for the first time. To design a PolSAR-tailored CL method, a diversity stimulation mechanism is constructed to replace the random sampling of ordinary CL methods so as to collect the diversified training data. This improvement can effectively narrow the application gap between optical and PolSAR images. After collecting the training datasets, the PCLNet includes two other parts, i.e., unsupervised pre-training and classifier fine-tuning. Among the former, the construction of memory bank effectively addresses the optimization difficulty, and the momentum-based update of auxiliary encoder significantly improves the consistency of the learning process. Therefore, supported by the results of unsupervised pre-training, high-precision few-shot PolSAR classification can be achieved by keeping the main encoder and fine-tuning a linear classifier. Numerous experiments are carried out on two widely-used benchmark datasets, and the experimental results exhibit the validity of PCLNet for both few-shot and full-supervised PolSAR classification compared with several popular methods. 
\par Compared with traditional CNN-based methods, the dataset collection and unsupervised pre-training of PCLNet undoubtedly needs more time for training. However, it achieves the utilization of massive unlabeled PolSAR data, and performs high-precision PolSAR classification with only a small amount of human annotations. We believe that more effective and appropriate pretext tasks may have potential to further improve the performance. More importantly, this work opens the door for future researches on unsupervised representation learning and few-shot, even zero-shot PolSAR image classification. Not only that, incorporating other application scenarios, like fine-grained classification, semantic segmentation and object detection, is our future interest.

\bibliographystyle{IEEEtran}
\bibliography{IEEEabrv,refference}
%\balance
%\vspace{-10 mm}
\begin{IEEEbiography}[{\includegraphics[width=1in,height=1.25in,clip,keepaspectratio]{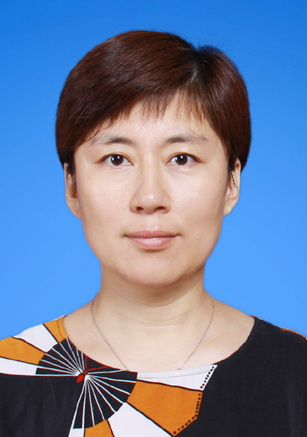}}]{Lamei Zhang} (M'07, SM’19) received the B.S., M.Sc., and Ph.D. degrees in information and communication engineering from Harbin Institute of Technology, Harbin, China, in 2004, 2006, and 2010, respectively. Currently, she is an associate professor with the Department of Information Engineering, Harbin Institute of Technology. She serves as the Secretary of IEEE Harbin GRSS Chapter. She was a Visiting Scholar in Department of Geological Sciences in University of Manitoba, Winnipeg, MB, Canada from 2014 to 2015. Her research interests include remote sensing images processing, information extraction and intelligent interpretation of high-resolution SAR, polarimetric SAR, and polarimetric SAR interferometry.
\end{IEEEbiography}
%\vspace{-10 mm}
\begin{IEEEbiography}[{\includegraphics[width=1in,height=1.25in,clip,keepaspectratio]{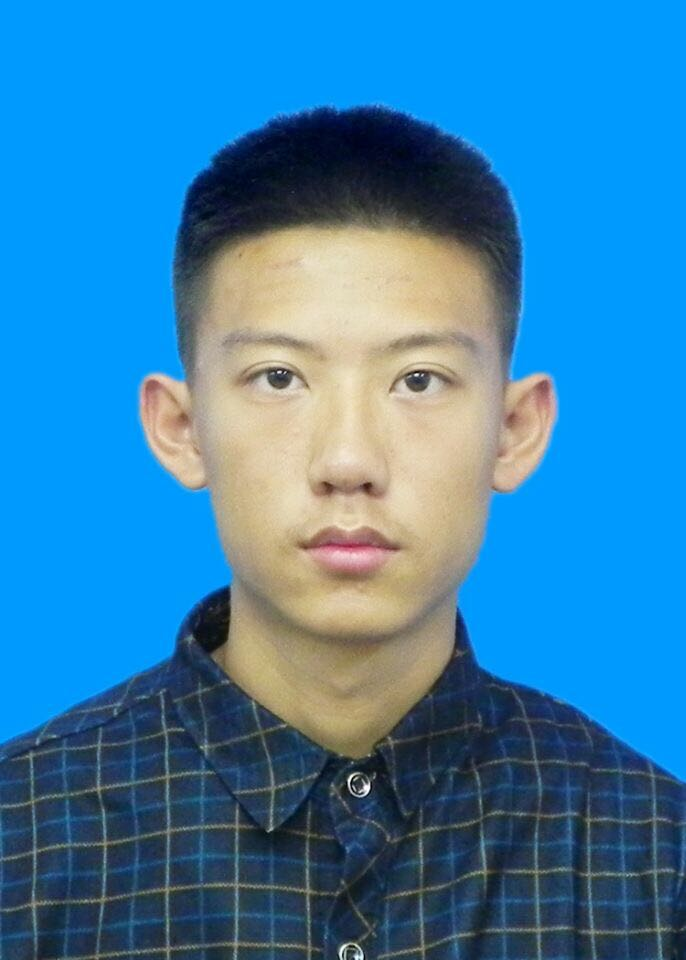}}]{Siyu Zhang} received the B.S. degree from electronic and information engineering, Dalian Maritime University, Dalian, China, in 2019. He is currently pursuing the M.S. degree with information and communication engineering, Harbin Institute of Technology, Harbin, China. His current research interests include PolSAR image interpretation and machine learning.
\end{IEEEbiography}
%\vspace{-10 mm}
\begin{IEEEbiography}[{\includegraphics[width=1in,height=1.25in,clip,keepaspectratio]{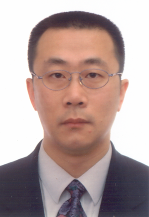}}]{Bin Zou} (M'04) received the B.S. degree in electronic engineering from Harbin Institute of Technology, Harbin, China, in 1990, the M.Sc. degree in space studies from the International Space University, Strasbourg, France, in 1998, and the Ph.D. degree in information and communication engineering from Harbin Institute of Technology, in 2001. From 1990 to 2000, he was with the Department of Space Electro-Optic Engineering, Harbin Institute of Technology. From 2003 to 2004, he was a Visiting Scholar with the Department of Geological Sciences, University of Manitoba, Winnipeg, MB, Canada. He is currently a Professor and Vice Head with the Department of Information Engineering, Harbin Institute of Technology. His research interests include SAR image processing, polarimetric SAR, and polarimetric SAR interferometry.
\end{IEEEbiography}
%\vspace{-10 mm}
\begin{IEEEbiography}[{\includegraphics[width=1in,height=1.25in,clip,keepaspectratio]{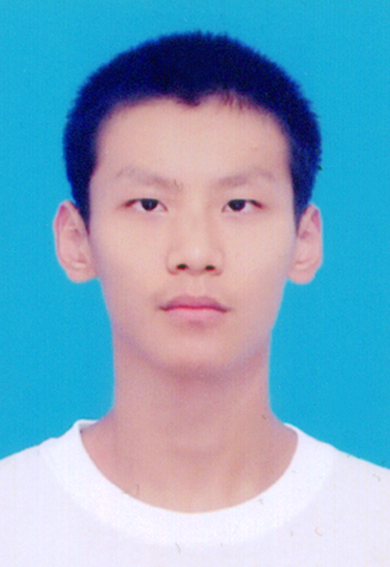}}]{Hongwei Dong} received the M.S. degree from College of Science, China Agricultural University, Beijing, China, in 2018. He is currently pursuing the Ph.D. degree in Harbin Institute of Technology, Harbin, China. His current research interests include applied mathematics, optimization methods and machine learning.
\end{IEEEbiography}
\end{document}